\pdfoutput=1
\documentclass{article}

\usepackage[preprint]{neurips_2023}

\usepackage[utf8]{inputenc} 
\usepackage[T1]{fontenc}    
\usepackage{hyperref}       
\usepackage{url}            
\usepackage{booktabs}       
\usepackage{amsfonts}       
\usepackage{nicefrac}       
\usepackage{microtype}      
\usepackage[table,xcdraw]{xcolor}         

\usepackage{subcaption}
\usepackage{wrapfig}
\usepackage{multirow}
\usepackage{makecell}
\usepackage{xspace}
\usepackage{enumitem}
\usepackage{amsmath}
\usepackage{listings}
\usepackage{algorithm}
\usepackage{algorithmic}
\usepackage{amsfonts,amssymb}
\usepackage{mathrsfs}
\usepackage{subcaption}
\usepackage{natbib}
\setcitestyle{numbers,square}
\usepackage{graphicx}       
\usepackage{cleveref}
\usepackage{tikz}
\usepackage{collcell}
\usepackage{etoolbox}
\usepackage{xfp}

\newcommand{\hhide}[1]{}
\newcommand{\hide}[1]{}
\newcommand{\model}[0]{\textsc{NCB}\xspace}
\newcommand{\vpara}[1]{\vspace{0.07in}\noindent\textbf{#1}\xspace} %

\def\zz#1{%
\newcommand{\MidNumber}{20}  
\newcommand{\MaxNumber}{39} 
\newcommand{\MinNumber}{0}   
\ifdim #1 pt > \MidNumber pt
            \pgfmathsetmacro{\PercentColor}{max(min(100.0*(#1 - \MidNumber)/(\MaxNumber-\MidNumber),100.0),0.00)} %
            \hspace{-0.33em}\colorbox{pink!\PercentColor!white}{#1}
        \else
            \pgfmathsetmacro{\PercentColor}{max(min(100.0*(\MidNumber - #1)/(\MidNumber-\MinNumber),100.0),0.00)} %
            \hspace{-0.33em}\colorbox{green!\PercentColor!white}{#1}
        \fi
}

\def\zzz#1{%
\newcommand{\MidNumber}{0}  
\newcommand{\MaxNumber}{10} 
\newcommand{\MinNumber}{-10}   
\ifdim #1 pt > \MidNumber pt
            \pgfmathsetmacro{\PercentColor}{max(min(100.0*(#1 - \MidNumber)/(\MaxNumber-\MidNumber),100.0),0.00)} %
            \hspace{-0.33em}\colorbox{green!\PercentColor!white}{#1}
        \else
            \pgfmathsetmacro{\PercentColor}{max(min(100.0*(\MidNumber - #1)/(\MidNumber-\MinNumber),100.0),0.00)} %
            \hspace{-0.33em}\colorbox{pink!\PercentColor!white}{#1}
        \fi
}

\title{NaturalCodeBench: Examining Coding Performance Mismatch on HumanEval and Natural User Prompts}

\author{
Shudan Zhang$^{12\dagger*}$, Hanlin Zhao$^{1*}$, Xiao Liu$^{12*}$, Qinkai Zheng$^{12*}$, \\
\bf{Zehan Qi$^{12\dagger}$, Xiaotao Gu$^{1}$, Xiaohan Zhang$^{1}$, Yuxiao Dong$^{2}$, Jie Tang$^{2}$}\\ \\
\textsuperscript{1}Zhipu.AI \quad
\textsuperscript{2}Tsinghua University \quad
\\ \\
{\includegraphics[height=3.5ex]{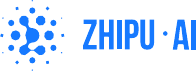}}
}

\begin{document}

\maketitle
\renewcommand{\thefootnote}{\fnsymbol{footnote}}
    \footnotetext[1]{SZ, HZ, XL, and QZ contributed equally. Emails: \{\texttt{zsd22@mails.tsinghua.edu.cn, hanlin.zhao@zhipuai.cn, shawliu9@gmail.com, qinkai.zheng1028@gmail.com}\}}
    \footnotetext[2]{Work done when SZ and ZQ interned at Zhipu AI.}
\renewcommand{\thefootnote}{\arabic{footnote}}

\vspace{-15mm}

\begin{figure}[h]
    \centering
    \includegraphics[width=1.0\textwidth]{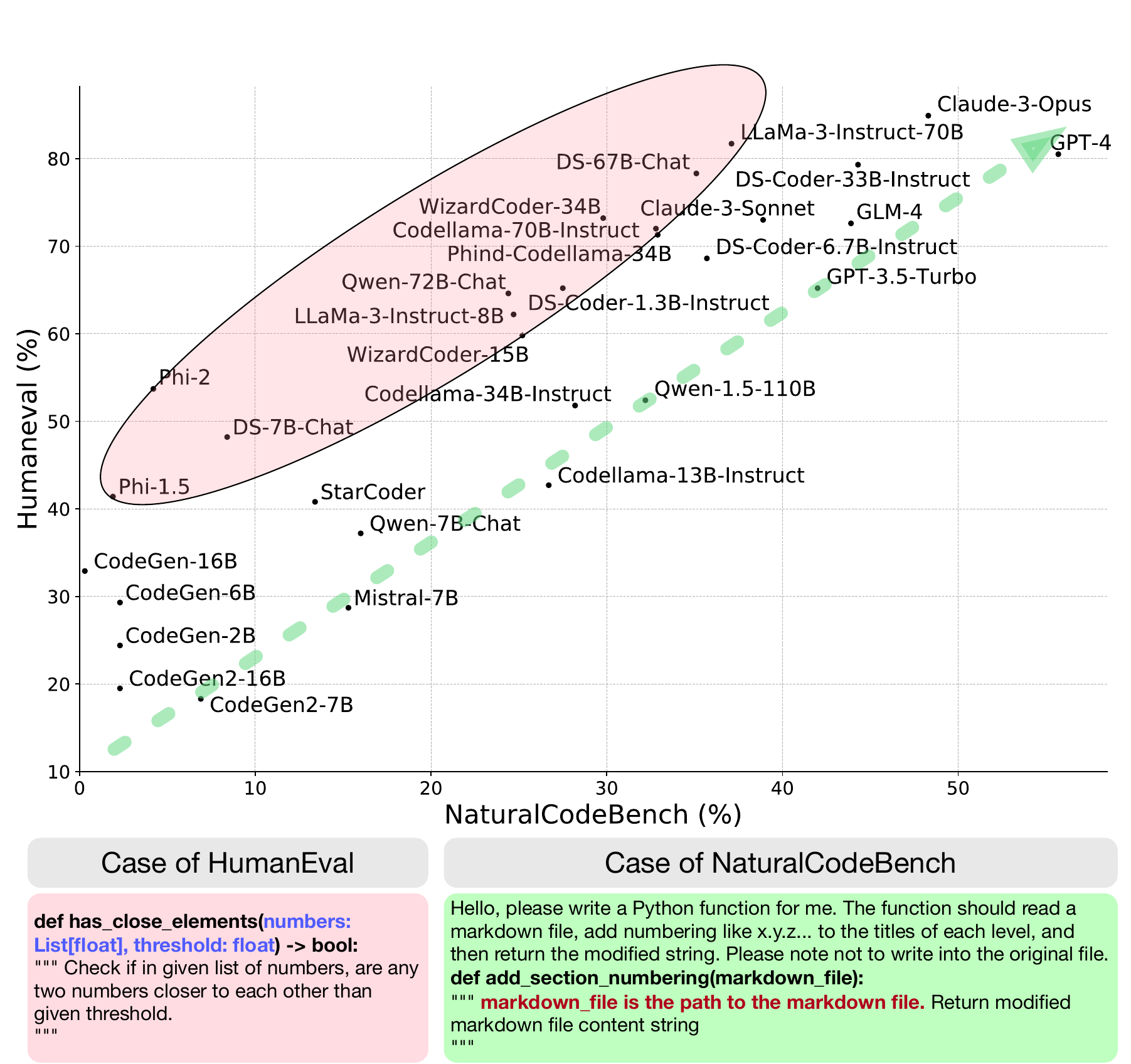}
    \caption{Comparison between HumanEval and \textsc{NaturalCodeBench}. (Upper) Performance plot of tested LLMs on both benchmarks. LLMs in red circle present relatively mismatched performances on two benchmarks. (Lower) Case study on coding tasks in HumanEval and \model. \model is grounded on natural prompts from real-world users and evaluated in an executable docker environment.}
    \label{fig:compare}
\end{figure}

\begin{abstract}
Large language models (LLMs) have manifested strong ability to generate codes for productive activities.
However, current benchmarks for code synthesis, such as HumanEval, MBPP, and DS-1000, are predominantly oriented towards introductory tasks on algorithm and data science, insufficiently satisfying challenging requirements prevalent in real-world coding.
To fill this gap, we propose \textsc{NaturalCodeBench} (\textsc{NCB}), a challenging code benchmark designed to mirror the complexity and variety of scenarios in real coding tasks. 
\model comprises 402 high-quality problems in Python and Java, meticulously selected from natural user queries from online coding services, covering 6 different domains. 
Noting the extraordinary difficulty in creating testing cases for real-world queries, we also introduce a semi-automated pipeline to enhance the efficiency of test case construction. 
Comparing with manual solutions, it achieves an efficiency increase of more than 4 times. 
Our systematic experiments on 39 LLMs find that performance gaps on \model between models with close HumanEval scores could still be significant, indicating a lack of focus on practical code synthesis scenarios or over-specified optimization on HumanEval. 
On the other hand, even the best-performing GPT-4 is still far from satisfying on \model.
The evaluation toolkit and development set are available at 
\url{https://github.com/THUDM/NaturalCodeBench}.

\end{abstract}

\section{Introduction}

Large language models (LLMs) pre-trained on extensive open code repositories~\cite{chen2021evaluating,openai2023gpt4,li2023starcoder,chowdhery2023palm} have demonstrated impressive performance on code synthesis and even achieve performance comparable to average human level in coding competitions~\cite{Li_2022}.
Unlike open text generation, which often underscores human preferences as noted by~\cite{NEURIPS2022_b1efde53}, code synthesis prioritizes accuracy and the fulfillment of user intent, essential for practical production and application.

As a result, evaluating code synthesis presents unique challenges in the era of LLMs.
Traditional evaluation metrics by token matching~\cite{papineni-etal-2002-bleu,lin-2004-rouge,popovic-2015-chrf} show a weak correlation with human judgement \cite{Evtikhiev_2023} and overlook functional correctness of the generated code \citealp{10.1145/3551349.3556903,10.1109/ICPC.2019.00034}.
Recently, execution-based evaluation has gained increasing popularity, where code generated by models is tested through unit tests to verify its functional correctness. 
It leads to the development of several benchmarks, including HumanEval \cite{chen2021evaluating}, MBPP \cite{austin2021program}, MBXP \cite{athiwaratkun2023multilingual}, CodeContests \cite{Li_2022}, and DS-1000 \cite{pmlr-v202-lai23b}.

Notwithstanding their commendable reliability and accuracy, these benchmarks fall short to sufficiently capture the wide range of needs and complexity found in real-world engineering applications.
They are primarily limited to well-defined coding problems in algorithm, program basics, or data science.
For example, as shown in Figure~\ref{fig:compare}, a problem from HumanEval~\cite{chen2021evaluating} tests the implementation of a basic function \texttt{has\_close\_elements} and takes floating-point arguments as inputs.
However, in practical applications, user engineering requirements can be much more complex and varied.
In Figure~\ref{fig:compare}, we showcase an example adapted from a real user query, where the user asks to read and parse XML files given certain tags.
Difficult and costly though it is, curating a benchmark composed of such problems is meaningful for evaluating the real user experience of LLM code synthesis.

\vpara{Contributions.}
In light of the challenge, we introduce \textsc{NaturalCodeBench} (\model), a challenging application-driven dataset for code synthesis evaluation. 
\model is dedicated to creating a reliable evaluation environment that is more aligned with real-world applications. 
We leverage an CodeGeeX \cite{10.1145/3580305.3599790} online services to collect real and diverse application-related user queries. 
After filtering and reprocessing, 402 high-quality Python and Java problems are compiled, covering 6 domains including software, front-end, system administration, and artificial intelligence, highlighting practical scenarios.
Beyond basic data structures like lists and numbers, the test inputs for \model problems include versatile file types and other complex structures, making it more challenging. 

The challenging nature of \model necessitates significant human labor in its annotation process
To improve construction efficiency, we tailor a semi-automated annotation pipeline to curate high-quality, testable, and useful queries with corresponding test cases. 
Specifically, we employ GPT-4 \cite{openai2023gpt4} to generate reference solutions followed by manual correction. 
Subsequently, GPT-4, guided by the problem descriptions and reference solutions, generates multiple test cases, which are also refined with manual correction, for each problem. 
Consequently, the annotators are only required to correct any errors, substantially reducing the time and manpower required.
Comparative experiments reveal that our semi-automated pipeline can quadruple the construction speed of the evaluation framework, as evidenced by tests involving programming experts with or without the pipeline.


Based on \model, we conduct extensive experiments on a variety range of LLMs, encompassing 39 APIs or open models.
The results indicate that although certain LLMs demonstrate comparable performance on established benchmarks like HumanEval, they exhibit significant performance disparities when evaluated using \model. 
It suggests that there may be inadequate focus on optimizing LLMs for practical coding applications, or have conducted over-specified optimization on HumanEval-style problems.
More importantly, even the best-performing GPT-4 only reaches about a pass rate of 53\%, demonstrating a large room for LLMs to improve their coding skills to face real-world coding challenges.

To facilitate community research, we pack up the whole \model testing environment into a docker image and make its development set publicly available.
To sum up our contributions:

\begin{itemize}[leftmargin=1.5em,itemsep=0pt,parsep=0.2em,topsep=0.1em,partopsep=0.0em]
\item[$\bullet$] We propose \textsc{NaturalCodeBench}, a benchmark that aligns with real-world applications, comprising 402 problems in Python and Java across 6 domains. We open source 140 problems (70 Python, 70 Java) as the development set of \model for research purposes, but keep the 262 problems of the test set closed to avoid contamination.
\item[$\bullet$] We introduce a semi-automated pipeline for the construction of code synthesis benchmarks, which significantly reduces time and manpower costs without compromising the quality of test cases. Comparative experiments reveal that our
semi-automated pipeline can quadruple the construction speed of the evaluation framework
\item[$\bullet$] We systematically benchmark the code generation capabilities of 39 LLMs using \model. Besides quantitative evaluation, we carry out a deep insight into the present stage of development in LLMs for code generation, and outline potential pathways for future progress.
\end{itemize}

\begin{figure*}[t]
    \centering
    \includegraphics[width=1\textwidth]{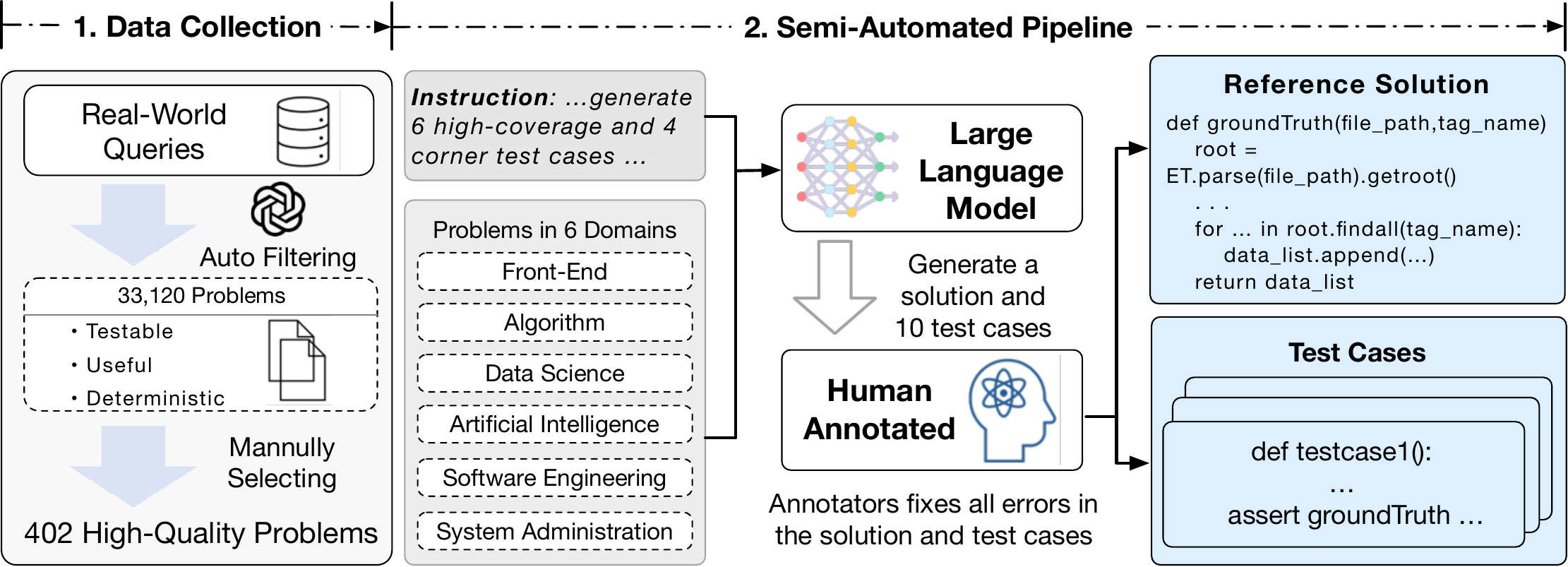}
    \caption{Overview of \textsc{NaturalCodeBench}. 1) Data Collection: collecting real-world queries from coding online services and selecting high-quality problems from the queries by GPT-3.5 and human annotators. 2) Semi-Automated Pipeline: improving efficiency of constructing evaluation framework by generating a solution and test cases with LLMs and then having them corrected by human annotators.}
    \label{fig:overview}
\end{figure*}

\section{Benchmark Construction}
The overview of \model is shown in Figure~\ref{fig:overview}. The pipeline of constructing \model consists of four steps: 1) collecting and filtering high-quality problems from online services (Section \ref{sec:select}) 2) constructing a complete evaluation framework through a semi-automated pipeline (Section \ref{sec:semi}) 3) designing prompts to align different models (Section \ref{sec:align}) 4) translating all problems and instructions to produce bilingual versions (Section \ref{sec:translation}).

\subsection{Problem Selection}
\label{sec:select}
\vpara{Collecting Real-World Problems.} 
To establish a meaningful and practical benchmark, we centered on collecting real-world code problems frequently encountered by users. To achieve this, the seed problems of \model are cleaned from the queries in coding online services. A part of users have granted permission for their data to be utilized exclusively for research purposes. We have strictly adhered to this directive by collecting only the relevant data from these consenting users and have implemented robust de-identification measures to eliminate any possibility of information leakage. We collect a varied collection of queries, spanning multiple programming languages, problem types, and levels of complexity. This diversity ensures that our benchmark accurately reflects a broad range of code issues users encountering in practice. We specifically concentrated on queries related to Python and Java, chosen for their widespread use in different domains.

\vpara{Filtering Testable Problems.} While it's possible to source inexhaustible queries from online services, many of these queries posed by users are either of low value or challenging to test the solution of these queries. For instance, some users may only seek basic clarifications on a built-in function, while others may not clearly articulate their objectives. To sieve out unsuitable queries for our testing, we've implemented a two-step filtering process. Initially, we employ GPT-3.5 to filter out low-quality queries, which saves on labour. This is achieved by adding specific criteria in the instruction, instructing GPT-3.5 to abandon those problems that cannot meet all specified requirements. These criteria are as follows: 1) Each query must involve at least one task, where the user requests the model's assistance in solving one or more problems. 2) Each query should be associated with several input-output pairs, ensuring that a given input correspond to a singular, definitive output. 3) The query must not contain any elements of randomness or uncertainty. The specifics of the instruction are detailed in (Appendix~\ref{sec:instruction}). Following this automated pre-screening, we conduct a manual review to further refine the selection, adhering to the outlined criteria. This process yields a final set of 201 unique Python and 201 unique Java problems. It is noteworthy that over 80\% of the initial queries failed to meet our stringent requirements.

\subsection{Semi-automated Pipeline}
\label{sec:semi}
In this section, we will introduce our semi-automated pipeline. To generate structurally complex and accurate test cases by GPT-4, it is first necessary to determine the arguments and return values of functions, as well as the names of objects. Therefore, a completely accurate reference solution is required initially. We generate a solution using GPT-4, then manually correct all errors. After this, based on the problem description and the reference solution, we instruct GPT-4 to generate multiple test cases. These are then reviewed by programming experts who correct errors and supplement any deficiencies in the generated test cases.

\vpara{Generating and Rewriting Reference Solution.}
GPT-4 is instructed to generate a solution for each problem in \model. It is important to note that while GPT-4 is highly capable, it is not infallible. Therefore, each solution generated by GPT-4 is meticulously examined by expert programmers to ensure correctness. In cases where the generated code contains errors, the expert programmers rewrite the code to rectify these issues. This process ensures the quality of the reference solutions. Even though we did not use the reference solution in \model for evaluation, we provided them to facilitate the generation of test cases and future research.

\vpara{Build High-Coverage and Corner Evaluation.}
We employ GPT-4 to generate evaluation codes for each problem. We construct a prompt using 1) the description of the problem for GPT-4 to inspect; 2) the reference solution to demonstrate the names and formats in the code; 3) an instruction to encourage GPT-4 to come up with effective test cases. Specifically, each prompt start with an instruction that ask GPT-4 to produce ten test cases based on the description of problem and the reference solution. Then, we present both the description of problem and its reference solution. We finalize the prompt with a initial segment of the evaluation code to assist GPT-4 in accurately generating the desired code format.
Our objective is to harness GPT-4's advanced comprehension and analytical abilities to learn valid format in the code and essential functionalities of the reference solution to enable the generation of superior test cases that are adept at uncovering latent errors in code. 

A complete and effective test should seek to identify potential bugs at different locations in the code, while also finding inputs that might trigger errors in the code. High coverage ensures that each test case examines more code and branches, thereby facilitating the discovery of concealed errors. Meanwhile, it is often observed that corner values in a problem's input are most prone to trigger code errors. Consequently, our instruction will cause some of the test cases generated by GPT-4 to have higher coverage, while the other part will be some corner values contained in the problem, so as to obtain more effective test cases.

Subsequently, expert programmers review and correct any test cases with formatting and answer errors. To ensure that the final evaluation framework is error-free.

\subsection{Alignment Between Different Models}
\label{sec:align}
In contrast to the problem format in Humaneval, the majority of problems in our benchmark are composed in natural language by actual users. Consequently, there is no predetermined naming convention for functions or classes created by models. This divergence can lead to inconsistencies between the names generated by LLMs and those referenced in test cases. To address this issue of name misalignment, we present a representative test case that includes the designated function or class name and its usage within the test. We then instruct the LLMs to adhere to the naming convention specified in the provided test case when generating solutions. It is important to note that the test cases utilized for solution generation are not employed in subsequent testing phases. The details of the instruction is showed in Appendix~\ref{sec:instruction}.

\subsection{Building Bilingual Benchmark}
\label{sec:translation}
The majority of the questions we collected from online services are in Chinese, which is not fair for the LLMs that are primarily designed for English. Therefore, we translate all the problems, resulting in both Chinese and English versions.

\begin{table*}[t]
\footnotesize
\setlength{\tabcolsep}{1pt}
\renewcommand\arraystretch{1.2}
\centering
\resizebox{\linewidth}{!}{
\begin{tabular}{@{}lccccccc@{}}
\toprule
\multirow{2}{*}{Benchmark}          & \multicolumn{5}{c}{Instruction Information}                                                  & \multicolumn{2}{c}{Evaluation} \\ \cmidrule(l){2-6} \cmidrule(l){7-8} 
                                    & \#Problem    & Domain               & \#Data Type & \#Word        & Source                   & \#Test Case   & Method                \\ \midrule
Humaneval \cite{chen2021evaluating} & 164          & Algorithm            & 5           & 23.0          & Hand-Written             & 7.7           & Test-Case             \\
MBPP \cite{austin2021program}       & 974          & Program Basics       & 5           & 15.7          & Hand-Written             & 3.0           & Test-Case             \\
DS-1000 \cite{pmlr-v202-lai23b}     & 1,000        & Data Sci.            & 6           & 140.0         & StackOverflow            & 1.6           & Test-Case + SFC.      \\
APPS \cite{hendrycksapps2021}       & 10,000       & Algorithm            & 5           & 293.2         & Competitions             & 13.2          & Test-Case             \\
Humaneval+ \cite{liu2023code}       & 164          & Algorithm            & 5           & 23.0          & Hand-Written             & 764.1         & Augmented Test Cases  \\ \midrule
\textbf{NaturalCodeBench}           & \textbf{402} & \textbf{Application} & \textbf{6}  & \textbf{78.3} & \textbf{Online Services} & \textbf{9.3}  & \textbf{Test-Case}    \\ \bottomrule
\end{tabular}
}
\caption{Comparison between \textsc{NaturalCodeBench} and other benchmarks for code generation. }
\label{tab:compare}
\end{table*}

\section{Dataset Statistics}

We provide more detailed statistics in Table~\ref{tab:statistics-ncb}. \model comprises a total of 402 problems collected from online services, with 201 problems in Python and 201 in Java, spanning across 6 domains: Database, Artificial Intelligence, Data Science, Algorithm and Data Structure, Front-End, Software Engineering, and System Administration. This diversity also leads to complex input data types in \model, which are classified into 9 categories: number (int/float/boolean), string, list (array), dict, tensor (matrix), data frame (table), plain text file, image, and special format file. The first four are the most common and simplest data types. Since a boolean can be represented by 1 and 0, we consider it as a type of number. Matrix and list are two similar types of data, but they are categorized separately due to differences in their usage scenarios. 
Due to the current popularity of deep learning, tensor has become a very common data format. Therefore, we have designated a separate category for tensor and have included matrix within this category. The last three are all file types, differentiated by their processing methods. The content of a plain text file is text and can be directly read. Figures require processing of each pixel value. A special format file refers to files that require specific methods for processing, such as PDF and DOCX.

Each problem within the dataset has been carefully curated with a set of test cases to assess the correctness of solutions. On average, there are 9.3 test cases associated with each problem. These cases are strategically designed, with about 60\% focused on enhancing statement and branch coverage, and the remaining 40\% dedicated to evaluating the robustness of solutions against corner values. The average word count for each problem in the \model is 78.3.

\vpara{Compared with Other Benchmark.} 
Table~\ref{tab:compare} compares \model to other benchmarks. It is noteworthy that our benchmark offers a substantial supplement to current benchmarks in terms of both problem and data types. Unlike Humaneval and MBPP, which consist of 96.9\% and 89.5\% algorithmic and basic programming problems respectively, our benchmark features a more balanced distribution across each domain. 

In addition, \model include more data types. Furthermore, \model focuses on assessing the model's ability to handle multiple file formats, a type of data that is both very commonly used in daily life and relatively challenging to process. We note that the problems involving files have fewer test cases, since GPT-4 still struggles to fully generate various types of file . This is also more challenging for human annotators to design compared to simpler data types.

On the other hand, \model is also limited by its size due to the high costs of problems collection and the construction of the evaluation framework. We are continuously working on expanding our benchmark.


\begin{table*}[t]
\footnotesize
\setlength{\tabcolsep}{13pt}
\centering
\begin{tabular}{@{}l|ccc|ccc@{}}
\toprule
           & \multicolumn{3}{c|}{\#Problems} & \multicolumn{3}{c}{Avg. \#Test Cases} \\ \cmidrule(l){2-7} 
Dataset    & Test     & Dev     & Total    & Test       & Dev       & Total      \\ \midrule
Software   & 88       & 44      & 132      & 9.7        & 8.6       & 9.3        \\
Data Sci.  & 68       & 32      & 100      & 9.6        & 8.6       & 9.3        \\
Algorithm  & 73       & 22      & 95       & 9.5        & 8.8       & 9.3        \\
Sys. Admin. & 22       & 17      & 33       & 9.6        & 8.5       & 9.1        \\
AI. System & 13       & 15      & 28       & 9.6        & 9.1       & 9.3        \\
Front-End  & 3        & 11      & 14       & 10.0       & 8.7       & 9.0        \\ \midrule
\textbf{Total/Avg.} & \textbf{262}      & \textbf{140}     & \textbf{402}      & \textbf{9.6 }       & \textbf{8.7}       & \textbf{9.3}        \\ \bottomrule
\end{tabular}
\caption{Detailed statistics of \textsc{NaturalCodeBench}.}
\label{tab:statistics-ncb}
\end{table*}

\section{Experiments}
\begin{table*}[t!]
\footnotesize
\setlength{\tabcolsep}{2pt}
\renewcommand\arraystretch{0.6}
\centering
\resizebox{0.93\textwidth}{!}{
\begin{tabular}{@{}l|c|ccc|ccc|cc|cc|c@{}}
\toprule
\textbf{Model}                        & \multirow{2}{*}{\textbf{Size}} & \multicolumn{3}{c|}{\textbf{NCB (zh)}}            & \multicolumn{3}{c|}{\textbf{NCB (en)}}            & \multicolumn{2}{c|}{\textbf{NCB Total}} & \multicolumn{2}{c|}{\textbf{HumanEval}} & \multirow{2}{*}{\textbf{$\Delta\text{Rank}$}} \\ 
                                                 &                                & \textbf{Python} & \textbf{Java} & \textbf{Total} & \textbf{Python} & \textbf{Java} & \textbf{Total} &         \textbf{Score}  & \textbf{Rank} & \textbf{Score}  & \textbf{Rank}  &                      \\ \midrule
\multicolumn{13}{c}{API LLMs} \\ \midrule
\textbf{GPT-4} \cite{openai2023gpt4}                                    & N/A                            & 53.4            & 51.1          & 52.3           & 55.7            & 51.1          & 53.4           & 52.8 & \zz{1}  &  80.5   & \zz{5} & \zzz{4}                     \\
\textbf{GPT-4-Turbo-0125} \cite{openai2023gpt4}                     & N/A                            & 51.4            & 58.6          & 55.0           & 48.6            & 51.4          & 50.0           & 52.5  & \zz{2}   & 87.2     & \zz{1} & \zzz{-1}                  \\
\textbf{GPT-4-Turbo-1106} \cite{openai2023gpt4}                         & N/A                            & 47.3            & 51.9          & 49.6           & 51.9            & 55.0          & 53.5           & 51.5  & \zz{3}   & 81.7     & \zz{3} & \zzz{0}                  \\
\textbf{GPT-3.5-Turbo} \cite{chatgpt}                           & N/A                            & 39.7            & 38.9          & 39.3           & 42.0            & 42.0          & 42.0           & 40.7   & \zz{8}    & 65.2   & \zz{18} & \zzz{10}                  \\ \midrule
\textbf{Claude-3-Opus} \cite{anthropic2023claude}                               & N/A                            & 45.0            & 50.4          & 47.7           & 48.9            & 48.9          & 48.9           & 48.3   & \zz{4}     & 84.9        & \zz{2} & \zzz{-2}           \\
\textbf{Claude-3-Sonnet} \cite{anthropic2023claude}                               & N/A                            & 44.6            & 35.5          & 40.1           & 40.5            & 35.1          & 37.8          & 38.9      & \zz{9}           & 73.0    & \zz{11} & \zzz{2}      \\
\textbf{Claude-3-Haiku} \cite{anthropic2023claude}                               & N/A                            & 41.3            & 35.9          & 38.6           & 36.9            & 30.5          & 33.7           & 36.2     & \zz{11}                & 75.9   & \zz{9} & \zzz{-2}   \\
\textbf{Claude-2.1} \cite{Claude-2}                              & N/A                            & 33.6            & 32.8          & 33.2           & 34.4            & 36.6          & 35.5           & 34.4     & \zz{13}            & 71.2       & \zz{16} & \zzz{3}    \\ \midrule
\textbf{ChatGLM-4} \cite{zeng2023glm130b,du-etal-2022-glm}                               & N/A                            & 43.5            & 45.3          & 44.4           & 41.5            & 45.3          & 43.4           & 43.9          & \zz{5}      & 72.6        & \zz{12} & \zzz{7}    \\ \midrule
\textbf{Gemini-1.5-Pro} \cite{google2024gemini}                               & N/A                            & 41.5            & 43.1          & 42.3           & 45.0            & 39.7          & 42.3           & 42.3     & \zz{7}          & 71.9        & \zz{14} & \zzz{7}     \\ \midrule
\textbf{CodeGeeX3} \cite{10.1145/3580305.3599790}                               & N/A                            & 29.0            & 29.0          & 29.0           & 36.6            & 32.8          & 34.7           & 31.9       & \zz{18}      & 69.5 & \zz{17} & \zzz{-1}               \\ \midrule
\multicolumn{13}{c}{Open LLMs} \\ \midrule
\multirow{3}{*}{\textbf{Deepseek-Coder-Instruct} \cite{guo2024deepseekcoder}}  & 33B                            & 44.3            & 38.9          & 41.6           & 44.3            & 44.3          & 44.3           & 43.0       & \zz{6}   & 79.3        & \zz{6} & \zzz{0}          \\
                                                  & 6.7B                           & 38.9            & 29.8          & 34.4           & 35.9            & 35.9          & 35.9           & 35.1 & \zz{12}  &78.6 & \zz{7} & \zzz{-5}                        \\
                                                  & 1.3B                           & 18.3            & 24.4          & 21.4           & 27.5            & 25.2          & 26.4           & 23.9  & \zz{22}  & 65.2    & \zz{19} & \zzz{-3}                    \\ \midrule
\multirow{2}{*}{\textbf{Llama-3-Instruct} \cite{llama3modelcard}}           & 70B                            & 39.1            & 34.4          & 36.7           & 35.4            & 39.7          & 37.5           & 37.1      & \zz{10}          & 81.7   & \zz{4} & \zzz{-6}         \\	
& 8B                             & 35.9             & 21.5          & 28.7            & 19.7             & 21.7          & 20.7           & 24.7      & \zz{21}          & 62.2   & \zz{21} & \zzz{0}        \\ \midrule		   
\multirow{2}{*}{\textbf{Deepseek-Chat} \cite{deepseek-llm}}           & 67B                            & 35.9            & 28.2          & 32.1           & 35.1            & 33.6          & 34.4           & 33.2      & \zz{14}             & 78.3     & \zz{8} & \zzz{-6}    \\
                                                  & 7B                             & 3.8             & 12.2          & 8.0            & 8.4             & 19.1          & 13.8           & 10.9        & \zz{30} &48.2           & \zz{26} & \zzz{-4}         \\ \midrule
\multirow{4}{*}{\textbf{Codellama-Instruct} \cite{roziere2023code}}      & 70B                            & 35.1            & 32.1          & 33.6           & 32.8            & 30.5          & 31.7           & 32.6        & \zz{15}         & 72.0    & \zz{13} & \zzz{-2}       \\
                                                  & 34B                            & 23.7            & 17.6          & 20.7           & 28.2            & 17.6          & 22.9           & 21.8     & \zz{24}   & 51.8     & \zz{25} & \zzz{1}               \\
                                                  & 13B                            & 20.6            & 16.8          & 18.7           & 26.7            & 19.1          & 22.9           & 20.8     & \zz{25}              & 42.7    & \zz{26} & \zzz{1}     \\
                                                  & 7B                             & 16.8            & 17.6          & 17.2           & 21.4            & 17.6          & 19.5           & 18.4    & \zz{26}                & 34.8   & \zz{31} & \zzz{5}     \\ \midrule
\textbf{Phind-Codellama} \cite{phind}                          & 34B                            & 34.4            & 29.0          & 31.7           & 33.6            & 32.1          & 32.9           & 32.3   & \zz{16}    & 71.3      & \zz{15} & \zzz{-1}               \\ \midrule
\textbf{Qwen-1.5} \cite{qwen}                          & 110B                            & 35.4            & 28.2          & 31.8           & 38.5            & 26.7          & 32.6           & 32.2      & \zz{17}    & 52.4  & \zz{24} & \zzz{7}               \\ \midrule
\multirow{2}{*}{\textbf{Qwen-Chat} \cite{bai2023qwen}}               & 72B                            & 28.2            & 29.8          & 29.0           & 24.4            & 29.0          & 26.7           & 27.9       & \zz{19}            & 64.6     & \zz{20} & \zzz{1}    \\
                                                  & 7B                             & 11.5            & 13.0          & 12.3           & 16.0            & 11.5          & 13.8           & 13.0      & \zz{28}           & 37.2 & \zz{30} & \zzz{2}          \\ \midrule
\multirow{2}{*}{\textbf{WizardCoder} \cite{luo2023wizardcoder}}             & 34B                            & 24.4            & 22.9          & 23.7           & 29.8            & 22.1          & 26.0           & 24.8        & \zz{20}              & 73.2  & \zz{10} & \zzz{-10}    \\
                                                  & 15B                            & 29.0            & 17.6          & 23.3           & 25.2            & 19.1          & 22.2           & 22.7   & \zz{23}            & 59.8    & \zz{22} & \zzz{-1}         \\ \midrule
\textbf{StarCoder} \cite{li2023starcoder}                                & 15.5B                          & 13.0            & 13.0          & 13.0           & 16.8            & 9.9           & 13.4           & 13.2      & \zz{27}               & 40.8  & \zz{29} & \zzz{2}      \\ \midrule
\textbf{Mistral-Instruct} \cite{jiang2023mistral}                         & 7B                             & 7.6             & 9.9           & 8.8            & 11.5            & 19.1          & 15.3           & 12.0          & \zz{29}              & 28.7   & \zz{34} & \zzz{5} \\ \midrule
\multirow{4}{*}{\textbf{CodeGen2} \cite{nijkamp2023codegen2}}                & 16B                            & 0.8             & 11.5          & 6.2            & 2.3             & 13.0          & 7.7            & 6.9       & \zz{31}                & 19.5   & \zz{36} & \zzz{5}   \\
                                                  & 7B                             & 2.3             & 5.3           & 3.8            & 6.9             & 5.3           & 6.1            & 5.0       & \zz{32}  & 18.3      & \zz{37} & \zzz{5}              \\
                                                  & 3.7B                           & 0.0             & 0.0           & 0.0            & 0.0             & 3.1           & 1.6            & 0.8  & \zz{38}  & 15.9       & \zz{38} & \zzz{0}                  \\
                                                  & 1B                             & 0.0             & 0.0           & 0.0            & 0.0             & 0.0           & 0.0            & 0.0   & \zz{39}               & 11.0  & \zz{39} & \zzz{0}         \\ \midrule
\multirow{2}{*}{\textbf{Phi} \cite{li2023textbooks}}                     & 2.7B                           & 5.3             & 3.1           & 4.2            & 3.1             & 5.3           & 4.2            & 4.2       & \zz{33}            & 53.7  & \zz{23} & \zzz{-10}        \\
                                                  & 1.3B                           & 0.0             & 0.8           & 0.4            & 3.8             & 0.0           & 1.9            & 1.2     & \zz{37}    & 41.4      & \zz{28} & \zzz{-9}              \\ \midrule
\multirow{3}{*}{\textbf{CodeGen} \cite{nijkamp2023codegen}}                 & 16B                            & 0.8             & 5.3           & 3.1            & 0.3             & 9.2           & 4.8            & 3.9         & \zz{34}         & 32.9       & \zz{32} & \zzz{-2}    \\
                                                  & 6B                             & 0.0             & 0.0           & 0.0            & 2.3             & 3.8           & 3.1            & 1.5     & \zz{35}        & 29.3       & \zz{33} & \zzz{-2}         \\
                                                  & 2B                             & 0.0             & 0.0           & 0.0            & 2.3             & 3.8           & 3.1            & 1.5     & \zz{36}             & 24.4    & \zz{35} & \zzz{-1}       \\ \bottomrule
\end{tabular}}
\caption{Evaluating LLMs on the test set of \textsc{NaturalCodeBench}. All results are pass@1 on greedy decoding. Dev set results are reported in Table~\ref{table:devset-result}. Compared to HumanEval~\cite{chen2021evaluating}, some LLMs present significant variations }
\label{table:main-result}
\vspace{-8mm}
\end{table*}

\subsection{Setup}
We conducted comprehensive evaluations of 33 popular state-of-the-art models. For proprietary models, our focus was on OpenAI's GPT-4-Turbo-0125, GPT-4-Turbo-1106, GPT-4, GPT-3.5-Turbo, Anthropic's Claude-2, ZhipuAI's CodeGeeX3. In the case of open-source models, we performed evaluations using the vLLM \cite{kwon2023efficient} and FastChat \cite{zheng2023judging} framework. Our evaluation primarily utilizes pass@k \cite{chen2021evaluating} as the metric to accurately assess the functional correctness of code generated by these models. For k equal to 1, we employ greedy-search decoding. For random sampling, we demonstrate the best pass@k results of the best-performing models with each LLM family for each $k \in \{10, 50\}$, where the sampling temperature is set to 0.2 and topp to 0.9.

Our semi-automated pipeline is capable of reducing the time required for benchmark construction without compromising the quality of test cases. This paper primarily focuses on evaluating the efficiency of benchmark construction and the quality of test cases. Specifically, we adopt code coverage \cite{10.1145/3338906.3340459}, a widely used metric for assessing the effectiveness of testing, as the criterion for evaluating the quality of test cases. We invite five programming experts, each tasked with constructing the same five problems. Initially, we ask each expert to manually write a standard solution and 5 test cases. Subsequently, for the same problems, they complete the writing of standard solutions and test cases using the semi-automated pipeline. As it is challenging to ensure identical test case coverage, we require that the test cases written under both methods should not have a code coverage of less than 80\%. Then, for the sake of convenient comparison, we calculate the scores for each construction method in a straightforward manner, which is outlined as follows:
$$
    Score = \frac{Line Cov. + Branch Cov.}{Time Cost} * 10
$$

\subsection{Results of LLMs}
Table~\ref{table:main-result} and Table~\ref{table:devset-result} shows the pass@1 results on the test set and dev set of \model, respectively. Considering the high consistency of results, we primarily analyze the results on the test set. As expected, OpenAI's GPT-4 achieves the highest score of 52.8\%. The performance of GPT-4-Turbo is very close to that of GPT-4, differing only by 1.3\% , with GPT-4-Turbo performing better in Java but showing a larger difference in Python. Among the open-source models, DeepSeek-Coder-33B-Instruct performs the best, reaching a score of 43.0\%. However, the 9.8\% score gap with GPT-4 remains significant. On the other hand, it surpasses the 40.7\% achieved by GPT-3.5, exceeding it by 2.3\%. In summary, the performance of state-of-the-art open-source models is now between GPT-3.5 and GPT-4, yet the majority of open-source models still do not match the performance of GPT-3.5. 

When compared to a perfect score of 100\%, it is observed that even the best-performing model, GPT-4, still falls significantly short. This is in contrast to its performance in HumanEval, where it has approached 90\%.

Comparing the performance of models in Chinese and English versions, it is evident that the vast majority of models perform better in English. This holds true even for the top models, GPT-4 and GPT-4-Turbo, which outperform their average scores in Chinese by 1.1\% and 3.9\%, respectively.

Furthermore, Table~\ref{table:main-result} systematically presents the performance of various open-source models at different scales. Models smaller than 10B scored between 0.0\% and 23.9\%, models between 10B and 30B scored between 3.9\% and 35.1\%, models between 30B and 60B scored between 21.8\% and 43.0\%, and models larger than 60B scored between 27.9\% and 33.2\%. It is evident that the scale of the model still has a significant impact on performance. Larger models generally outperform smaller models, indicating that increasing scale can indeed enhance a model's capabilities. However, this is not to say that scale is everything; more refined data and training strategies can also significantly impact a model's performance. Some smaller models, such as DeepSeek-Coder-6.7B-Instruct, can outperform those larger than 30B by approximately 2.8\% and those larger than 60B by approximately 1.9\%. 

Table~\ref{tab:passk} shows the pass@k results of best-performing LLMs with each LLM family on \model, where $k \in \{10, 50\}$. We found that under random sampling, the scores of some models increased significantly. For instance, Codellama-70B-Instruct, unlike its performance on pass@1, clearly outperformed GPT-3.5 on both Pass@10 and Pass@50.

We compared the Python scores on the test set of \model with the performances of models on Humaneval, as shown in the Figure~\ref{fig:compare}. Most models are located in the upper triangular area of the graph, with many models scoring high on Humaneval but exhibiting relatively lower performance on \model.

\begin{table*}[t]
\footnotesize
\setlength{\tabcolsep}{7pt}
\centering
\resizebox{\linewidth}{!}{
\begin{tabular}{@{}l|cccc|cccc@{}}
\toprule
                    & \multicolumn{4}{c|}{\textbf{Hand-Written}}                                              & \multicolumn{4}{c}{\textbf{Semi-Automated}}                                             \\ \midrule
                    & \textbf{Time Cost} & \textbf{Line} & \textbf{Branch} & \textbf{Score} & \textbf{Time Cost} & \textbf{Line} & \textbf{Branch} & \textbf{Score} \\ \midrule
\textbf{Expert\_1}  & 179.5              & 97.6                   & 95.9                     & 10.8           & 36.0               & 97.0                   & 96.9                     & 53.9           \\
\textbf{Expert\_2}  & 195.0              & 97.6                   & 95.0                     & 9.9            & 41.0               & 88.1                   & 91.7                     & 43.9           \\
\textbf{Expert\_3}  & 145.0              & 84.5                   & 84.0                     & 11.6           & 26.0               & 82.0                   & 85.0                     & 64.2           \\
\textbf{Expert\_4}  & 180.0              & 90.9                   & 100.0                    & 10.6           & 41.0               & 84.4                   & 91.7                     & 42.9           \\
\textbf{Expert\_5}  & 180.0              & 98.1                   & 83.3                     & 10.1           & 56.0               & 100.0                  & 100.0                    & 35.7           \\ \midrule
\textbf{Total/Avg.} & 175.9              & 93.7                   & 91.6                     & 10.5           & 40.0               & 90.3                   & 93.1                     & 48.1           \\ \bottomrule
\end{tabular}}
\caption{Test case construction comparison between by Semi-Automated Pipeline and Hand-Written}
\label{tab:semi-result}
\vspace{-4mm}
\end{table*}

\subsection{Performance mismatch on HumanEval and \model}
We show the rank orders of all tested LLMs in Table~\ref{table:main-result} with regard to HumanEval and \model, as well as the difference of rank orders.
We also plot the corresponding performances on two benchmarks to scatter diagram in Figure~\ref{fig:compare}.
Based on the table and figure, we have some interesting findings.

Performances of most LLMs on two benchmarks grow linearly proportional, and the differences of scores' rank order are around 0.
It demonstrates that \model can indeed reflect the coding abilities of LLMs as HumanEval does in most cases.

However, we observe that some model series, notably the Phi, Deepseek-Chat, and WizardCoder, consistently exhibit a propensity to achieve superior rankings on the Humaneval dataset as opposed to the \model across various scales, as shown in the Table ~\ref{table:main-result}. 
Additional model families, including CodeGen and Llama-3-Instruct, similarly display the trend, though to a reduced degree. 

There might be a few potential hypotheses for the observation.
First, as problems in \model are more difficult and derived from natural user prompts, compared to those in HumanEval, LLMs with poorer generalization and instruction-following capabilities tend to perform worse. 
We find in preliminary experiments that problems in \model cannot be properly solved by pre-trained base LLMs via mere in-context learning as HumanEval does, which indicates that to solve \model problems requires stronger alignment and generalizability than HumanEval needs.

Second, it is possible that training sets of some LLMs are over-specifiedly optimized for HumanEval-style problems.
On one hand, pre-training data of certain LLMs may be contaminated.
As GPT-4~\cite{openai2023gpt4} reported, 25\% of HumanEval has been contaminated in their pre-training corpus.
On the other hand, instruction fine-tuning dataset may also be polluted.
For example, Phi~\cite{li2023textbooks} reports a considerable amount of synthetic prompts resonating to some test samples in HumanEval.
In~\cite{yang2023rethinking}, the authors report leakage unidentifiable by n-gram overlap when using popular rephrasing techniques to create training sets.
The performance discrepancy between HumanEval and \model in our experiments is also an evidence of the potential contamination.

\subsection{Results of Semi-automated Construction}
In Table ~\ref{tab:semi-result}, we can observe that the coverage of hand-written test cases is almost identical to that of test cases constructed through a semi-automatic pipeline, yet the time required for the former significantly exceeds the time needed for constructing test cases via the semi-automatic pipeline. Specifically, test cases can be constructed via the semi-automated pipeline in just 40 minutes, whereas manual writing requires 175.9 minutes, a difference of more than 4x. Consequently, the scores obtained for test cases constructed using the semi-automated pipeline are far higher than those for manually written test cases, with an average difference of 37.6. In summary, constructing test cases through the semi-automatic framework can achieve significantly higher efficiency without substantial loss in quality compared to manual writing.

\section{Related Work}
\vpara{LLMs for code.} 
Significant advancements in LLMs (\citealp{10.5555/3295222.3295349}, \citealp{devlin-etal-2019-bert}, \citealp{NEURIPS2020_1457c0d6}) are transforming everyday life, particularly in the field of coding, driven by the vast amount of openly available codebases and the push to enhance productivity among developers. Code-specific LLMs have proven their ability to perform various tasks such as code generation (\citealp{chen2021evaluating}, \citealp{iyer-etal-2018-mapping}, \citealp{Li_2022}), program repair (\citealp{10.1109/ICSE48619.2023.00125}, \citealp{10.1145/3611643.3616271}, \citealp{10.1109/ICSE48619.2023.00129}, \citealp{10.1145/3540250.3549101}), automated testing (\citealp{10.1145/3597926.3598067}, \citealp{deng2023large}, \citealp{10.1109/ICSE48619.2023.00119}, \citealp{xia2024fuzz4all}, \citealp{yang2023whitebox}), code translation (\citealp{10.5555/3495724.3497454}, \citealp{roziere2022leveraging}) and code summarization (\citealp{10.1145/3551349.3559555}, \citealp{lu2021codexglue}). Notably, prominent LLMs including CODEX \cite{chen2021evaluating}, CodeGen \cite{nijkamp2023codegen}, INCODER \cite{fried2023incoder}, and PolyCoder \cite{10.1145/3520312.3534862} have been developed and rigorously tested, particularly in code generation. This area, often referred to as the ultimate goal in computer science research since the early days of AI in the 1950s, involves the model producing code snippets from natural language explanations of the required functionality. The landscape of code LLMs is currently experiencing a surge, with new models being introduced regularly. This includes both proprietary ones (\citealp{10.1016/j.jss.2023.111734}, \citealp{openai2023gpt4}) and open-source ones (\citealp{lin-2004-rouge}, \citealp{nijkamp2023codegen}, \citealp{touvron2023llama}, \citealp{li2023starcoder}, \citealp{anonymous2024wizardcoder}, \citealp{rozière2024code}), marking a trend of frequent releases in this domain.

\vpara{Code Synthesis Benchmarks.} 
As the capabilities of models advance, researchers are developing more challenging and versatile benchmarks for code generation. 
Initially, the earlier focus was on domain-specific languages \cite{10.5555/1864519.1864543}, while the subsequent effort launched a Text-to-SQL benchmark to evaluate the capacity for generating comprehensive SQL programs \cite{yu-etal-2018-spider}. 
A investigation \cite{10.1145/3196398.3196408} assesses the ability to compose brief yet broadly applicable Python snippets. 
More recent studies (\citealp{hendrycks2021measuring}, \citealp{Li_2022}) have tested models' proficiency in solving competitive programming challenges using Python. 
A leading and extensively researched benchmark in this domain is HumanEval \cite{chen2021evaluating}, which features 164 Python function signatures accompanied by docstrings and corresponding test cases for validating correctness. 
Additionally, each problem in HumanEval includes a reference solution. The MBPP \cite{austin2021program} dataset, another Python-centric collection, was developed by having participants contribute 974 programming challenges. 
Each challenge encompasses a problem description (i.e., docstring), a function signature, and three test cases. There are also benchmarks for other programming languages, such as HumanEval-X \cite{10.1145/3580305.3599790} for C++, JavaScript, and Go, CodeContests \cite{Li_2022} for C++ and Java, and MultiPL-E \cite{cassano2022multiple}, which expands HumanEval and MBPP to 18 languages. 

More recent efforts have introduced benchmarks that more closely mirror real-world coding scenarios that require interactive coding. 
For example, AgentBench~\cite{liu2023agentbench} introduces interactive tasks regarding unix shell and MySQL. 
SWE-Bench~\cite{jimenez2023swe} compiles GitHub issues, their associated codebases, and tests, to gauge LLMs' effectiveness in practical software engineering tasks.

\section{Conclusion}
We propose \textsc{NaturalCodeBench} for evaluating the code generating ability of LLMs. Our benchmark comprises a total of 402 problems selected from coding online services, and it supports automatic evaluation of code generated by LLMs. We have also proposed a semi-automated pipeline for efficiently constructing the entire benchmark, achieving an efficiency gain of more than 4x compared to manual construction. We hope that \model can provide a fair environment for the comparison between models, and our pipline can also provide inspiration to other complex tasks or domains where evaluation costs are high.

\section*{Limitations}
Here, we discuss several limitations of this work.

\vpara{To cover more domains.}
Although our problems are derived from real-world application scenarios, due to the difficulty of constructing accurate and efficient evaluation environments, we are unable to test some types of problems, such as those involving interface creation, web services, etc., which are also common problem types in actual applications. This results in some biases in our evaluation, which may affect the accuracy of the evaluation of certain models. We will leave these issues for future research.

\vpara{To reduce the cost.}
The semi-automated pipeline can significantly reduce the time and human resources required to construct an evaluation framework, but the cost of accessing OpenAI's API remains expensive, and it does not completely eliminate the use of human resources.

\bibliographystyle{abbrv}
\bibliography{ref}

\appendix
\section{Instructions in \textsc{NaturalCodeBench}}
\label{sec:instruction}
\begin{figure}[ht]
    \centering
    \includegraphics[width=0.48\textwidth]{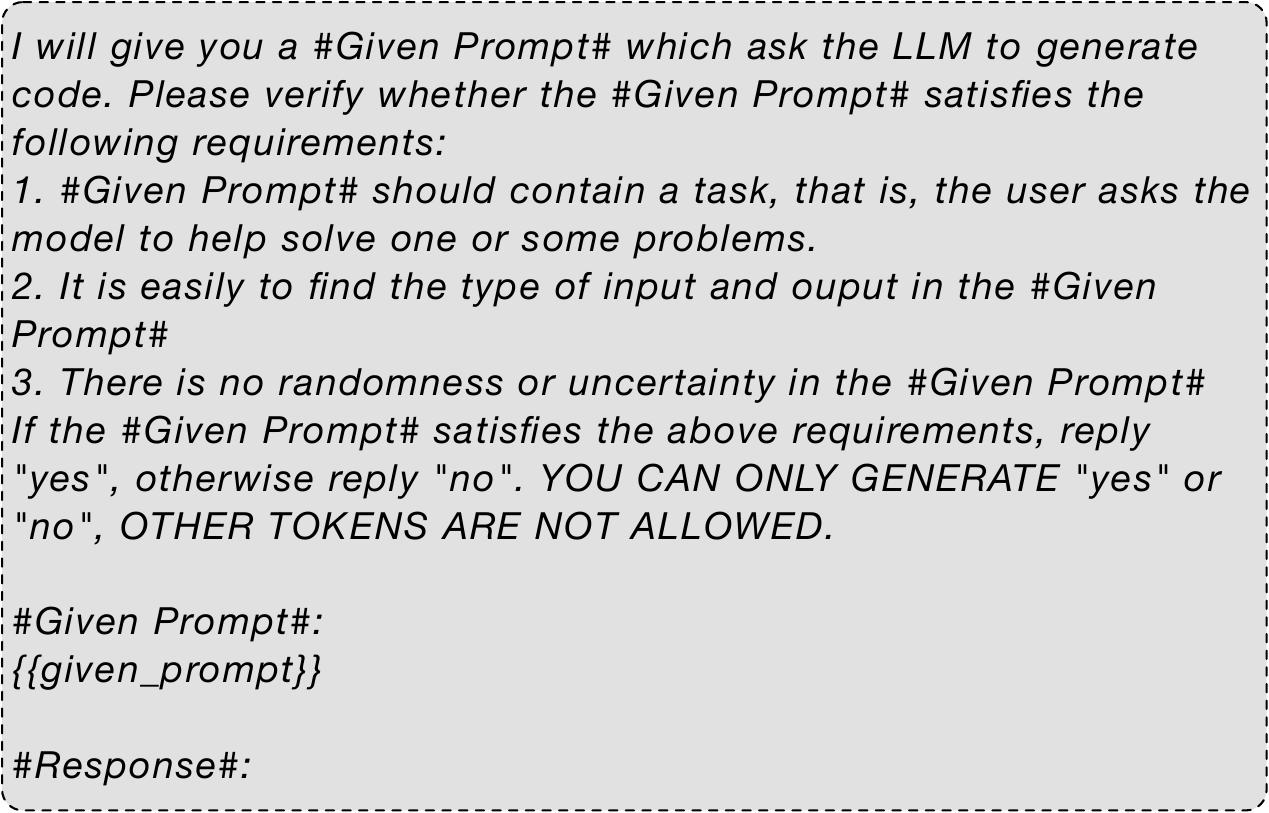}
    \caption{The instruction used to quickly filter out low-quality queries}
    \label{fig:testable}
\end{figure}

\begin{figure}[ht]
    \centering
    \includegraphics[width=0.48\textwidth]{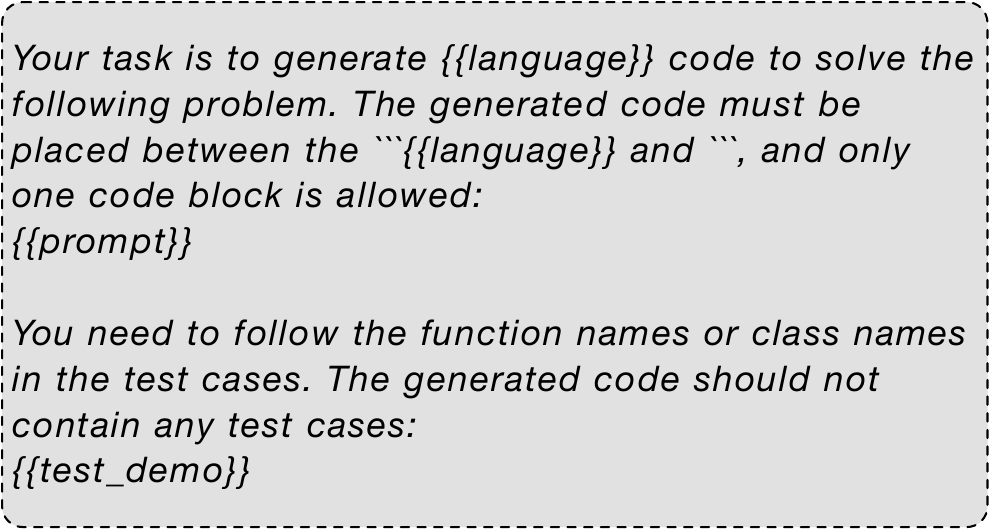}
    \caption{The instruction used to align the names of classes or functions generated by the LLMs with the names in the test cases.}
    \label{fig:alignment}
\end{figure}

\begin{figure*}[h]
    \centering
    \includegraphics[width=0.9\textwidth]{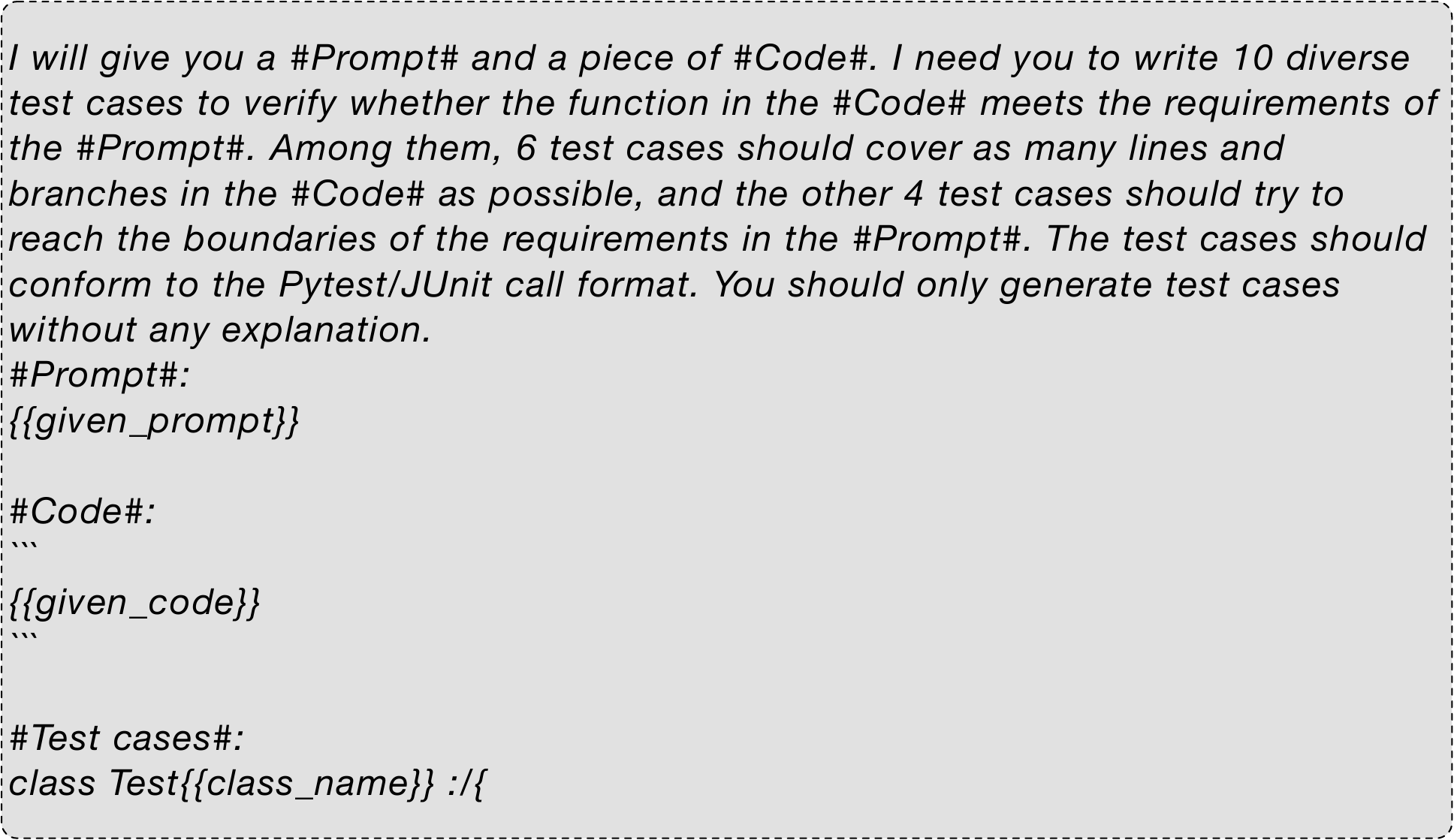}
    \caption{The insturciton used in Semi-automated Pipeline. Generating 6 test cases for high-coverage and 4 corner test cases.}
    \label{fig:testcases}
\end{figure*}

To enhance the efficiency of benchmark construction and reduce human labor costs, we utilized the extensive knowledge storage and natrual language understanding capabilities of LLMs during the benchmark construction process. Below are the details of the instructions used in the construction process:
\begin{itemize}[leftmargin=1.5em,itemsep=0pt,parsep=0.2em,topsep=0.1em,partopsep=0.0em]
\item[$\bullet$] Figure~\ref{fig:testable} shows the instruction we employed to swiftly filter out queries unsuitable for testing.
\item[$\bullet$] Figure~\ref{fig:testcases} shows how we instruct the GPT-4 to generate diverse and high-quality testcases.
\item[$\bullet$] Figure~\ref{fig:alignment} illustrates how we address the issue of misalignment between class or function names generated by the LLMs and the names in the test cases.
\end{itemize}

\section{Examples}
\subsection{Examples of Semi-Automated Pipeline}
In this section, we present two examples, one each for Python and Java, of semi-automated pipeline with one test case to illustrate how we construct test cases and rectify errors therein.

\begin{figure*}[h]
    \centering
    \includegraphics[width=0.86\textwidth]{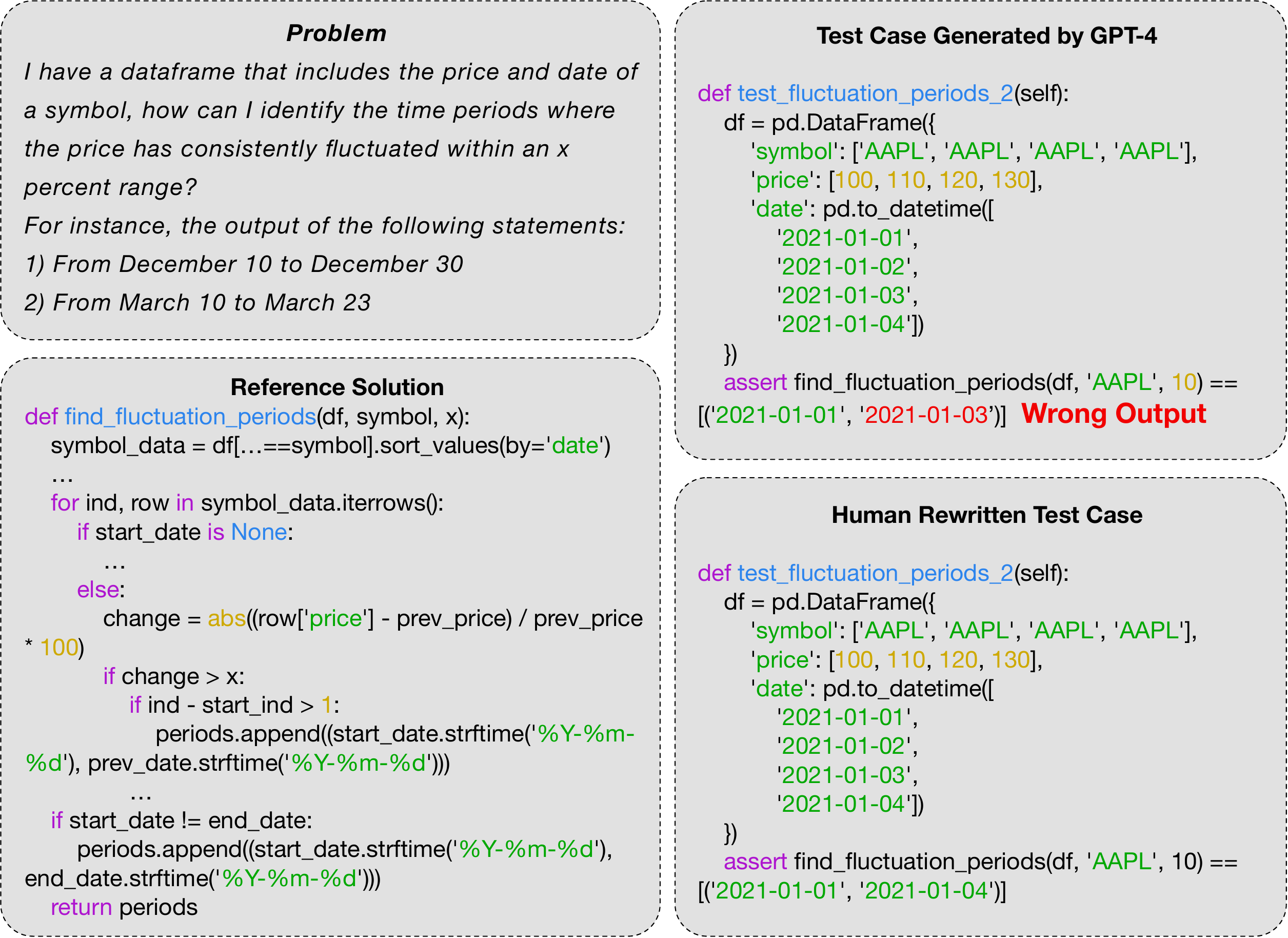}
    \caption{A Python example of semi-automate pipeline.}
    \label{fig:semi-python}
\end{figure*}

Figure~\ref{fig:semi-python} shows the Python example. Following the provision of problem description and reference solution, GPT-4 writes the majority of the test case, including the execution procedure and test case input. However, GPT-4 could not guarantee the correctness of each test case, resulting in the generation of erroneous expected outputs. At this point, our programming experts only needed to correct the incorrect expected outputs.

\begin{figure*}[h]
    \centering
    \includegraphics[width=0.86\textwidth]{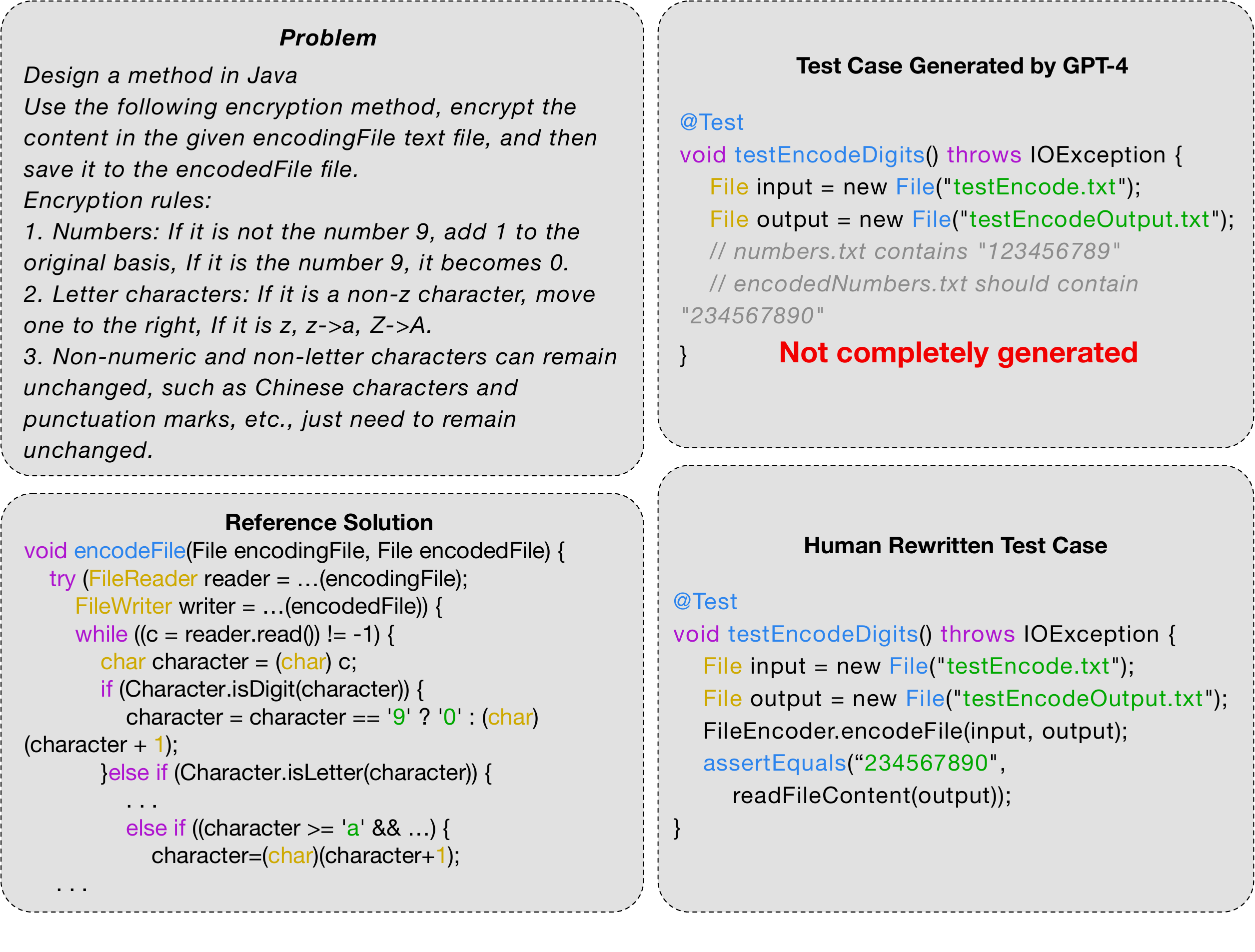}
    \caption{A Java example of semi-automate pipeline.}
    \label{fig:semi-java}
\end{figure*}

Figure~\ref{fig:semi-java} shows the Java exmaple. In this problem, where the input type involves more complex file formats, our semi-automatic pipeline is unable to directly generate the input files corresponding to each test case. Therefore, in this instance, our programming experts need to not only supplement the missing procedures in the test cases but also create an input file for each test case. However, GPT-4 has provided reference content for the input files in the comments, so our programming experts do not need to design the inputs themselves.

\subsection{Example Problems}
\begin{table*}[t]
\footnotesize
\setlength{\tabcolsep}{3pt}
\centering
\resizebox{\linewidth}{!}{
\begin{tabular}{@{}l|c|cccc|cccc@{}}
\toprule
                                                \textbf{Model} & \multicolumn{1}{c|}{\multirow{3}{*}{\textbf{Dataset}}} & \multicolumn{4}{c|}{\textbf{NCB(zh)}}                                                      & \multicolumn{4}{c}{\textbf{NCB(en)}}                                                      \\ \cmidrule(l){3-10} 
                                                      & \multicolumn{1}{c|}{}                         & \multicolumn{2}{c}{\textbf{Python}}         & \multicolumn{2}{c|}{\textbf{Java}}           & \multicolumn{2}{c}{\textbf{Python}}         & \multicolumn{2}{c}{\textbf{Java}}           \\ \cmidrule(l){3-10} 
                                                      & \multicolumn{1}{c|}{}                         & \textbf{Pass@10}     & \textbf{Pass@50}     & \textbf{Pass@10}     & \textbf{Pass@50}      & \textbf{Pass@10}     & \textbf{Pass@50}     & \textbf{Pass@10}     & \textbf{Pass@50}     \\ \midrule
\multirow{2}{*}{\textbf{GPT-4} \cite{openai2023gpt4}}                       & Test                                          & 62.4                 & 67.9                 & 64.6                 & 71.8                  & 65.3                 & 70.2                 & 62.7                 & 67.9                 \\
                                                      & Dev                                           & 53.3                 & 55.7                 & 69.2                 & 72.9                  & 51.8                 & 54.3                 & 62.0                 & 64.3                 \\ \midrule
\multirow{2}{*}{\textbf{GPT-3.5-Turbo} \cite{chatgpt}}               & Test                                          & 46.5                 & 48.9                 & 49.3                 & 56.5                  & 53.5                 & 55.7                 & 51.5                 & 57.3                 \\
                                                      & Dev                                           & 44.0                 & 47.7                 & 45.5                 & 51.4                  & 43.6                 & 47.1                 & 48.4                 & 50.0                 \\ \midrule
\multirow{2}{*}{\textbf{Deepseek-Coder-33B-Instruct} \cite{guo2024deepseekcoder}} & Test                                          & 55.7                 & 61.8                 & 48.0                 & 51.1                  & 56.6                 & 64.9                 & 52.8                 & 59.5                 \\
                                                      & Dev                                           & 48.1                 & 51.4                 & 46.8                 & 51.4                  & 46.5                 & 48.6                 & 46.7                 & 50.0                 \\ \midrule
\multirow{2}{*}{\textbf{Codellama-70B-Instruct} \cite{roziere2023code}}      & Test                                          & 49.6                 & 56.5                 & 52.7                 & 61.8                  & 51.0                 & 62.6                 & 48.2                 & 58.0                 \\
                                                      & Dev                                           & 47.5                 & 54.3                 & 53.9                 & 62.9                  & 47.6                 & 54.3                 & 50.5                 & 60.0                 \\ \midrule
\multirow{2}{*}{\textbf{Phind-Codellama-34B} \cite{phind}}         & Test                                          & 42.3                 & 46.6                 & 39.4                 & 45.8                  & 40.6                 & 43.5                 & 47.6                 & 56.5                 \\
                                                      & Dev                                           & 45.4                 & 50.0                 & 41.7                 & 45.7                  & 44.0                 & 45.7                 & 49.4                 & 51.4                 \\ \midrule
\multirow{2}{*}{\textbf{Deepseek-67B-Chat} \cite{deepseek-llm}}       & Test                                          & 44.3                 & 48.9                 & 40.8                 & 47.8                  & 47.3                 & 51.9                 & 40.9                 & 45.8                 \\
                                                      & Dev                                           & 42.3                 & 47.1                 & 44.5                 & 47.1                  & 37.9                 & 41.4                 & 43.6                 & 50.0                 \\ \midrule
\multirow{2}{*}{\textbf{Qwen-72B-Chat} \cite{bai2023qwen}}               & Test                                          & 34.9                 & 37.4                 & 36.5                 & 39.7                  & 32.7                 & 35.9                 & 36.5                 & 38.2                 \\
                                                      & Dev                                           & 43.4                 & 47.1                 & 31.4                 & 38.6                  & 41.0                 & 44.3                 & 31.5                 & 35.7                 \\ \midrule
\multirow{2}{*}{\textbf{StarCoder} \cite{li2023starcoder}}                   & Test                                          & 23.1 & 28.2 & 23.3 & 29.8 & 24.1 & 31.3 & 26.8 & 32.1 \\
                                                      & Dev                                           & 29 & 32.9 & 27.3 & 32.9 & 35.5 & 41.4 & 27.0 & 30.0 \\ \midrule
\multirow{2}{*}{\textbf{Mistral-7B-Instruct} \cite{jiang2023mistral}}         & Test                                          & 15.5                 & 18.3                 & 17.3                 & 20.6                  & 19.6                 & 22.9                 & 22.0                 & 24.4                 \\
                                                      & Dev                                           & 18.2                 & 21.4                 & 16.3                 & 20.0                  & 19.7                 & 24.3                 & 17.8                 & 21.4                 \\ \midrule
\multirow{2}{*}{\textbf{CodeGen2-16B} \cite{nijkamp2023codegen2}}                & Test                                          & 8.6 & 16.8 & 18.0 & 22.9 & 13.0 & 19.1 & 21.0 & 26.0 \\
                                                      & Dev                                           & 11.6 & 21.4 & 12.8 & 15.7 & 16.0 & 24.3 & 18.5 & 24.3 \\ \midrule
\multirow{2}{*}{\textbf{CodeGen-16B} \cite{nijkamp2023codegen}}                 & Test                                          & 4.6 & 9.2 & 13.3 & 18.3 & 9.9 & 15.3 & 17.5 & 21.4 \\
                                                      & Dev                                           & 10.7 & 17.1 & 15.6 & 18.6 & 16.1 & 22.9 & 17.4 & 21.4 \\ \midrule
\multirow{2}{*}{\textbf{Phi-2} \cite{li2023textbooks}}                       & Test                                          & 14.5                 & 21.4                 & 5.5                  & 7.6                   & 11.9                 & 19.8                 & 10.7                 & 14.5                 \\
                                                      & Dev                                           & 15.3                 & 27.1                 & 5.1                  & 7.1                   & 10.9                 & 18.6                 & 6.4                  & 7.1                  \\ \bottomrule
\end{tabular}}
\caption{Pass@k results of best-performing LLMs with each LLM family on NaturalCodeBench.}
\label{tab:passk}
\end{table*}
\begin{table*}[t!]
\footnotesize
\setlength{\tabcolsep}{4pt}
\renewcommand\arraystretch{0.96}
\centering
\resizebox{\linewidth}{!}{
\begin{tabular}{@{}l|c|ccc|ccc|c@{}}
\toprule
\textbf{Model}                        & \multirow{2}{*}{\textbf{Size}} & \multicolumn{3}{c|}{\textbf{NCB(zh)}}            & \multicolumn{3}{c|}{\textbf{NCB(en)}}            & \multirow{2}{*}{\textbf{Total}} \\ \cmidrule(lr){3-8}
                                                  &                                & \textbf{Python} & \textbf{Java} & \textbf{Total} & \textbf{Python} & \textbf{Java} & \textbf{Total} &                                 \\ \midrule
\multicolumn{9}{c}{API LLMs} \\ \midrule
\textbf{GPT-4} \cite{openai2023gpt4}                                   & N/A                            & 50.0            & 64.3          & 57.2           & 47.1            & 57.1          & 52.1           & 54.6                            \\
\textbf{GPT-4-Turbo-1106} \cite{openai2023gpt4}                        & N/A                            & 54.3            & 55.7          & 55.0           & 50.0            & 54.3          & 52.2           & 53.6                            \\
\textbf{GPT-4-Turbo-0125} \cite{openai2023gpt4}                        & N/A                            & 51.5            & 55.7          & 53.6           & 48.6            & 51.4          & 50.0           & 51.8                            \\
\textbf{GPT-3.5-Turbo} \cite{chatgpt}                           & N/A                            & 38.6            & 38.6          & 38.6           & 37.1            & 41.4          & 39.3           & 38.9                            \\ \midrule
\textbf{Claude-3-Opus} \cite{anthropic2023claude}                               & N/A                            & 46.4            & 44.3          & 45.3           & 50.0            & 47.1          & 48.6           & 47.0                           \\
\textbf{Claude-3-Haiku} \cite{anthropic2023claude}                               & N/A                            & 40.3            & 32.9          & 36.6           & 43.8            & 32.9          & 38.4           & 37.5                           \\
\textbf{Claude-3-Sonnet} \cite{anthropic2023claude}                               & N/A                            & 37.8            & 41.4          & 39.6           & 38.6            & 31.4          & 35.0           & 37.3                           \\
\textbf{Claude-2.1} \cite{Claude-2}                              & N/A                            & 41.4            & 37.1          & 39.3           & 35.7            & 35.7          & 35.7           & 37.5                            \\ \midrule
\textbf{ChatGLM-4} \cite{zeng2023glm130b,du-etal-2022-glm}                               & N/A                            & 42.9            & 47.1          & 45.0           & 44.3            & 42.9          & 43.6           & 44.3                            \\ \midrule
\textbf{Gemini-1.5-Pro} \cite{google2024gemini}                               & N/A                            & 44.3            & 35.7          & 40.0           & 48.6            & 34.3          & 41.4           & 40.7                            \\ \midrule
\textbf{CodeGeeX3} \cite{10.1145/3580305.3599790}                               & N/A                            & 40.0            & 25.7          & 32.9           & 35.7            & 25.7          & 30.7           & 31.8                            \\ \midrule
\multicolumn{9}{c}{Open LLMs} \\ \midrule
\multirow{3}{*}{\textbf{Deepseek-Coder-Instruct} \cite{guo2024deepseekcoder}}  & 33B                            & 41.4            & 40.0          & 40.7           & 35.7            & 41.4          & 38.6           & 39.6                            \\
                                                  & 6.7B                           & 34.3            & 40.0          & 37.2           & 34.4            & 40.0          & 37.2           & 37.2                            \\
                                                  & 1.3B                           & 22.9            & 21.4          & 22.2           & 20.0            & 27.1          & 23.6           & 22.9                            \\ \midrule
\multirow{2}{*}{\textbf{Llama-3-Instruct} \cite{llama3modelcard}}           & 70B                            & 42.9            & 37.1          & 40.0           & 37.1            & 41.4          & 39.3           & 39.6                            \\	
& 8B                             & 22.9            & 20.0          & 21.4            & 12.9             & 20.0          & 16.4           & 18.9                           \\ \midrule
\textbf{Phind-Codellama} \cite{phind}                          & 34B                            & 34.1            & 31.4          & 32.8           & 38.6            & 40.0          & 39.3           & 36.0                            \\ \midrule
\textbf{Qwen-1.5} \cite{qwen}                          & 110B                            & 35.7            & 30.0          & 32.9           & 37.1            & 35.7          & 36.4           & 34.6                            \\ \midrule
\multirow{4}{*}{\textbf{Codellama-Instruct} \cite{roziere2023code}}      & 70B                            & 30.0            & 30.0          & 30.0           & 35.7            & 35.7          & 35.7           & 32.9                            \\
                                                  & 34B                            & 14.3            & 25.7          & 20.0           & 25.7            & 25.7          & 25.7           & 22.9                            \\
                                                  & 13B                            & 21.4            & 20.0          & 20.7           & 22.9            & 20.0          & 21.5           & 21.1                            \\
                                                  & 7B                             & 25.7            & 14.3          & 20.0           & 18.6            & 17.1          & 17.9           & 18.9                            \\ \midrule
\multirow{2}{*}{\textbf{Deepseek-Chat} \cite{deepseek-llm}}           & 67B                            & 28.6            & 35.7          & 32.2           & 28.6            & 32.9          & 30.8           & 31.5                            \\
                                                  & 7B                             & 12.9            & 11.4          & 12.2           & 10.0            & 14.3          & 12.2           & 12.2                            \\ \midrule
\multirow{2}{*}{\textbf{WizardCoder} \cite{luo2023wizardcoder}}             & 34B                            & 31.4            & 31.4          & 31.4           & 30.0            & 31.4          & 30.7           & 31.1                            \\
                                                  & 15B                            & 30.0            & 24.3          & 27.2           & 31.4            & 24.3          & 27.9           & 27.5                            \\ \midrule
\multirow{2}{*}{\textbf{Qwen-Chat} \cite{bai2023qwen}}               & 72B                            & 35.7            & 24.3          & 30.0           & 34.3            & 25.7          & 30.0           & 30.0                            \\
                                                  & 7B                             & 10.0            & 12.9          & 11.5           & 20.0            & 15.7          & 17.9           & 14.7                            \\ \midrule
\textbf{StarCoder} \cite{li2023starcoder}                                & 15.5B                          & 17.1            & 15.7          & 16.4           & 21.4            & 15.7          & 18.6           & 17.5                            \\ \midrule
\textbf{Mistral-Instruct} \cite{jiang2023mistral}                         & 7B                             & 11.4            & 12.9          & 12.2           & 15.7            & 11.4          & 13.6           & 12.9                            \\ \midrule
\multirow{4}{*}{\textbf{CodeGen2} \cite{nijkamp2023codegen2}}                & 16B                            & 5.7             & 7.1           & 6.4            & 8.6             & 7.1           & 7.9            & 7.1                             \\
                                                  & 7B                             & 1.4             & 5.7           & 3.6            & 1.4             & 5.7           & 3.6            & 3.6                             \\
                                                  & 3.7B                           & 0.0             & 5.7           & 2.9            & 2.9             & 2.9           & 2.9            & 2.9                             \\
                                                  & 1B                             & 0.0             & 2.9           & 1.5            & 0.0             & 2.9           & 1.5            & 1.5                             \\ \midrule
\multirow{3}{*}{\textbf{CodeGen} \cite{nijkamp2023codegen}}                 & 16B                            & 1.4             & 5.7           & 3.6            & 7.1             & 8.6           & 8.6            & 5.7                             \\
                                                  & 6B                             & 2.9             & 2.9           & 2.9            & 4.3             & 7.1           & 5.7            & 4.3                             \\
                                                  & 2B                             & 0.0             & 2.9           & 1.5            & 2.9             & 5.7           & 4.3            & 2.9                             \\ \midrule
\multirow{2}{*}{\textbf{Phi} \cite{li2023textbooks}}                     & 2.7B                           & 4.3             & 4.3           & 4.3            & 5.7             & 4.3           & 5.0            & 4.7                             \\
                                                  & 1.3B                           & 1.4             & 2.9           & 2.2            & 5.7             & 4.3           & 5.0            & 3.6                             \\ \bottomrule
\end{tabular}}
\caption{Evaluating LLMs on the dev set of \textsc{NaturalCodeBench}. All results are pass@1 on greedy decoding.}
\label{table:devset-result}
\end{table*}
Here, we present an example problem and test cases for each of the 6 domains.

\begin{figure*}[h]
    \centering
    \includegraphics[width=0.95\textwidth]{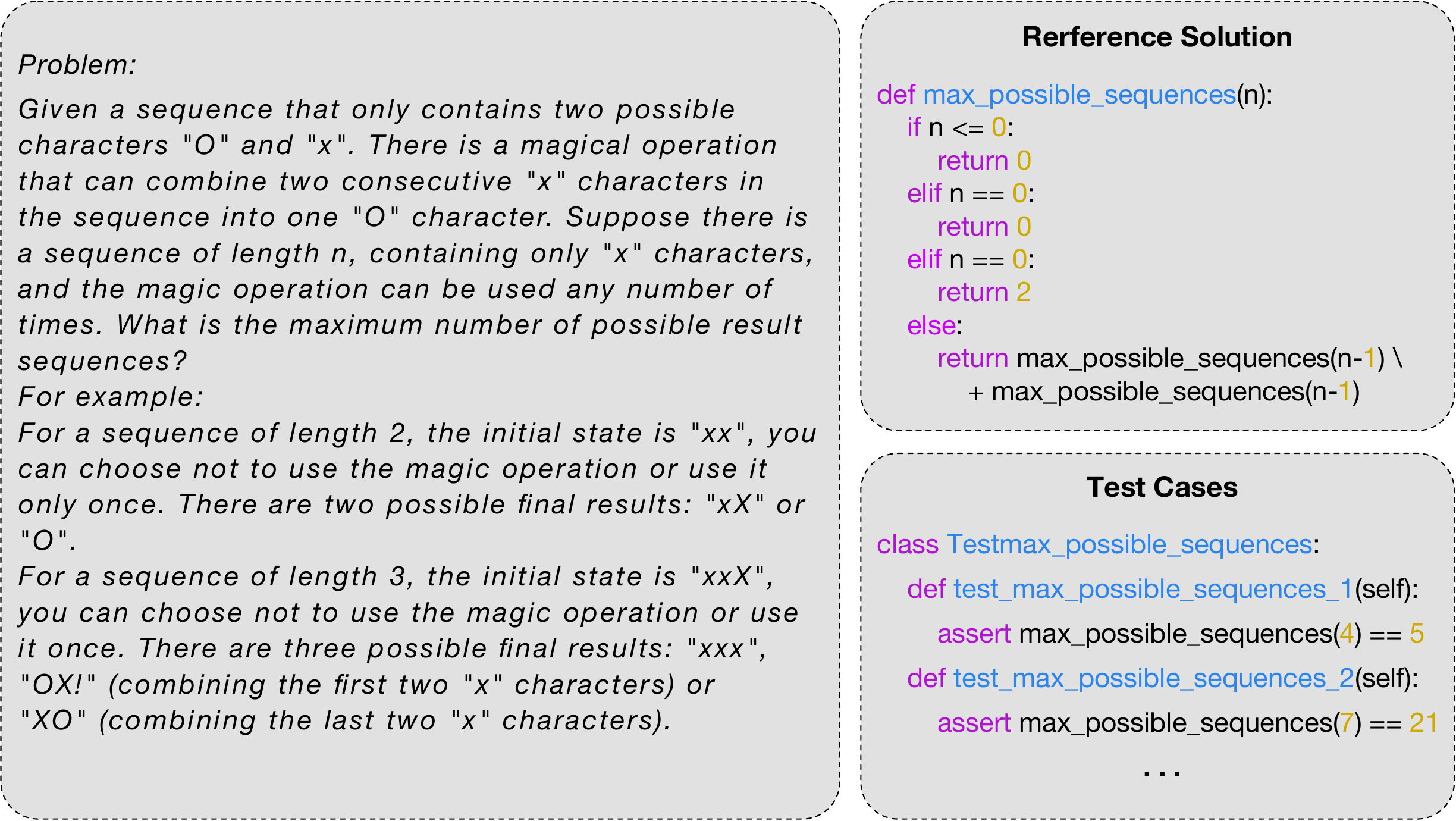}
    \caption{An example problem of Algorithm and Data Structure.}
    \label{fig:algorithm}
\end{figure*}

Figure~\ref{fig:algorithm} shows a problem of Algorithm and Data Structure, querying the pattern of a sequence transformation and the total number of all transformations.

\begin{figure*}[h]
    \centering
    \includegraphics[width=0.95\textwidth]{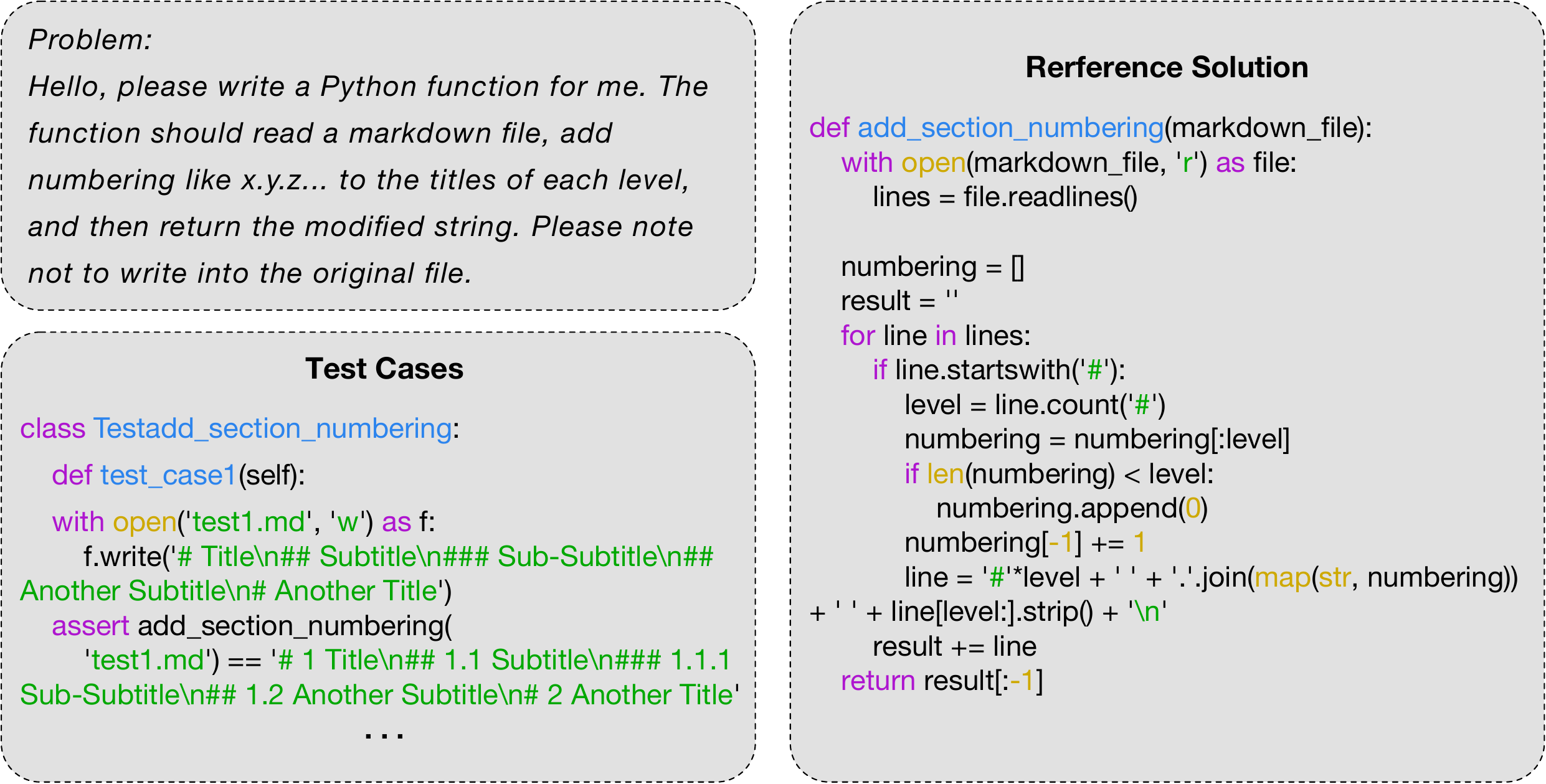}
    \caption{An example problem of Software Engineering.}
    \label{fig:software}
\end{figure*}

Figure~\ref{fig:software} illustrates an example problem in software engineering, requiring the addition of tags to different titles in a markdown file according to their levels.

\begin{figure*}[h]
    \centering
    \includegraphics[width=0.95\textwidth]{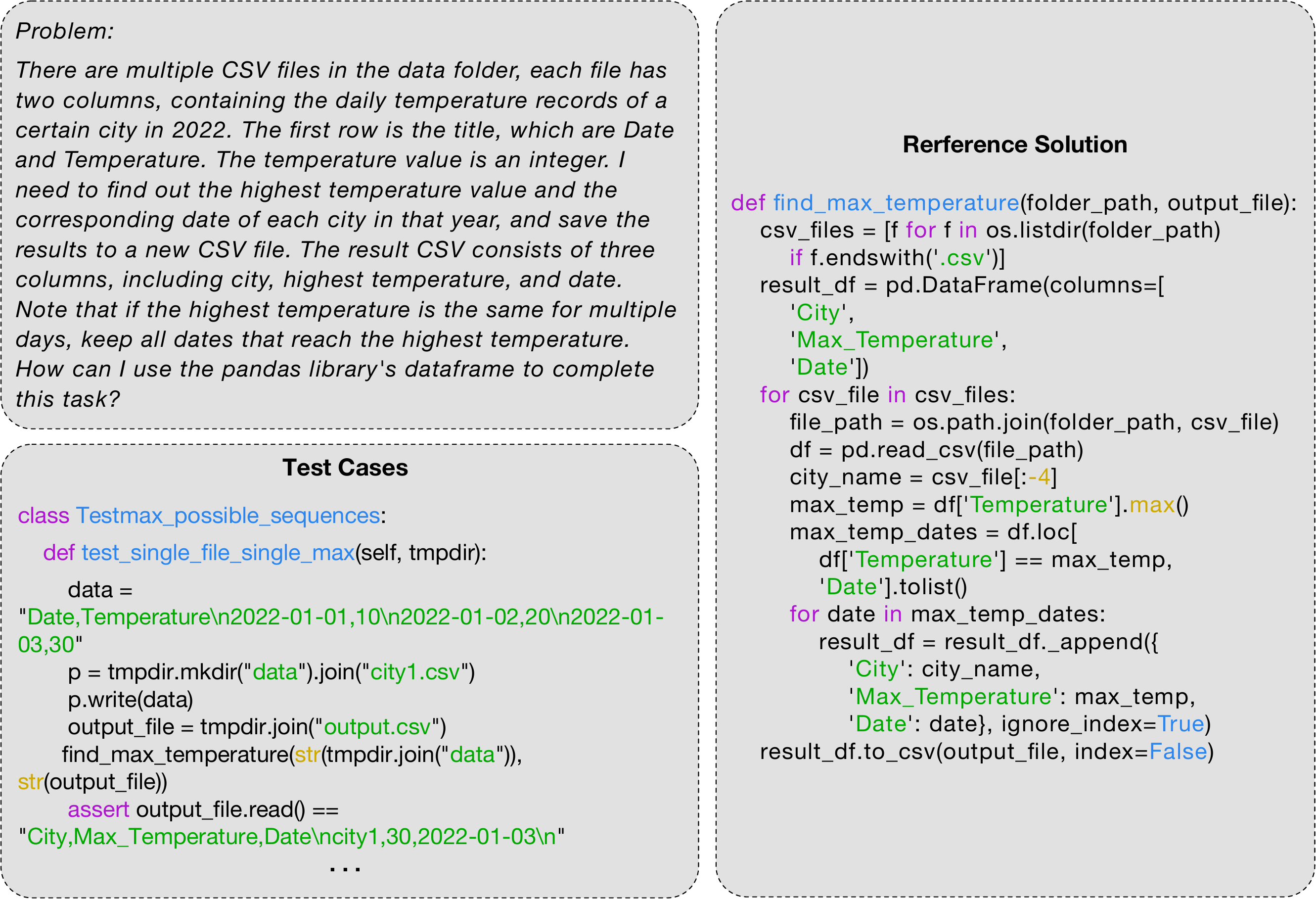}
    \caption{An example problem of Data Science.}
    \label{fig:data-science}
\end{figure*}

Figure~\ref{fig:data-science} presents an example problem in data science, asking to select the row with the highest temperature from the temperature CSV files of each city and write these rows into a new CSV file.

\begin{figure*}[h]
    \centering
    \includegraphics[width=0.95\textwidth]{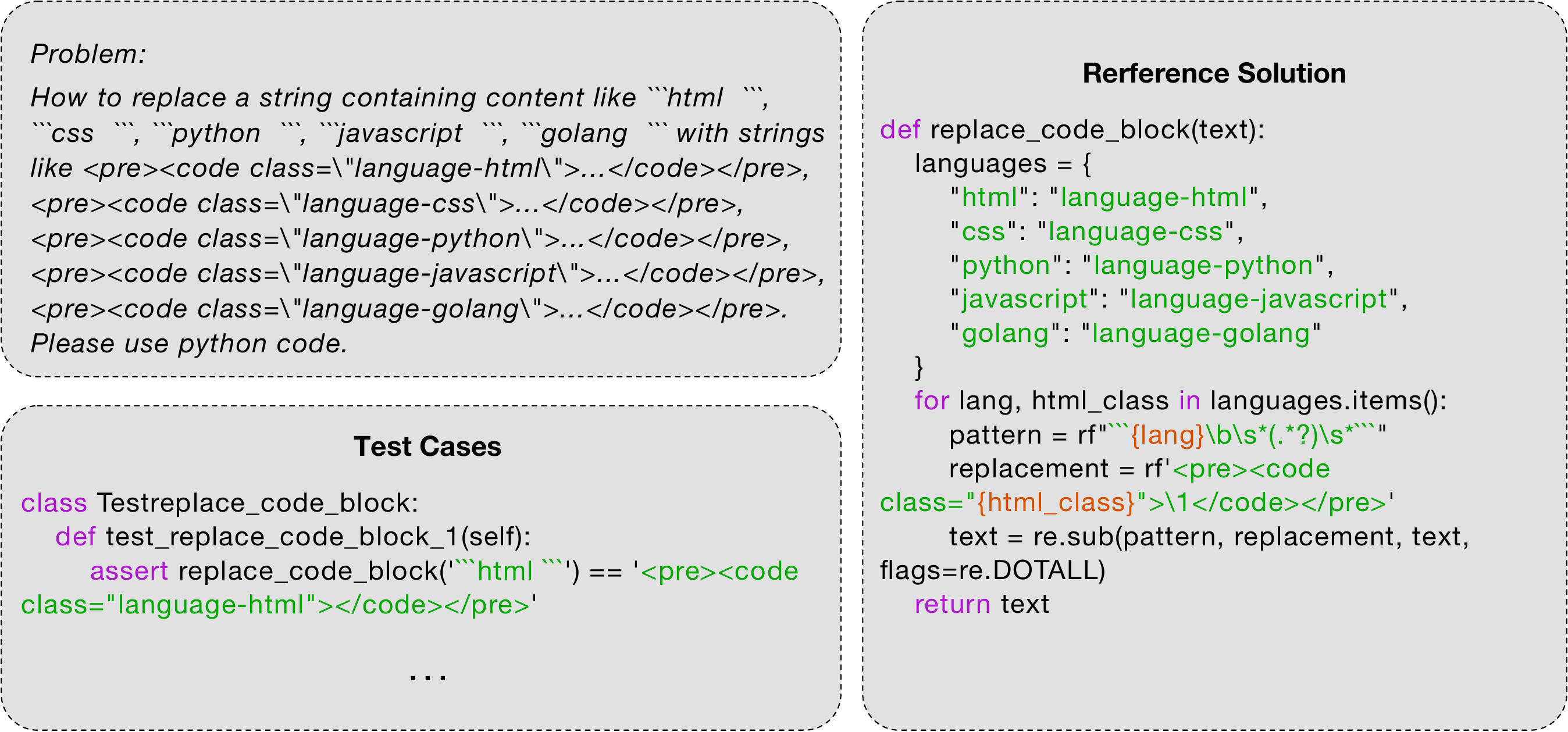}
    \caption{An example problem of Front-End.}
    \label{fig:front}
\end{figure*}

Figure~\ref{fig:front} depicts an example problem in front-end development, requiring the replacement of given special tags within a string with specific HTML formats.

\begin{figure*}[h]
    \centering
    \includegraphics[width=0.95\textwidth]{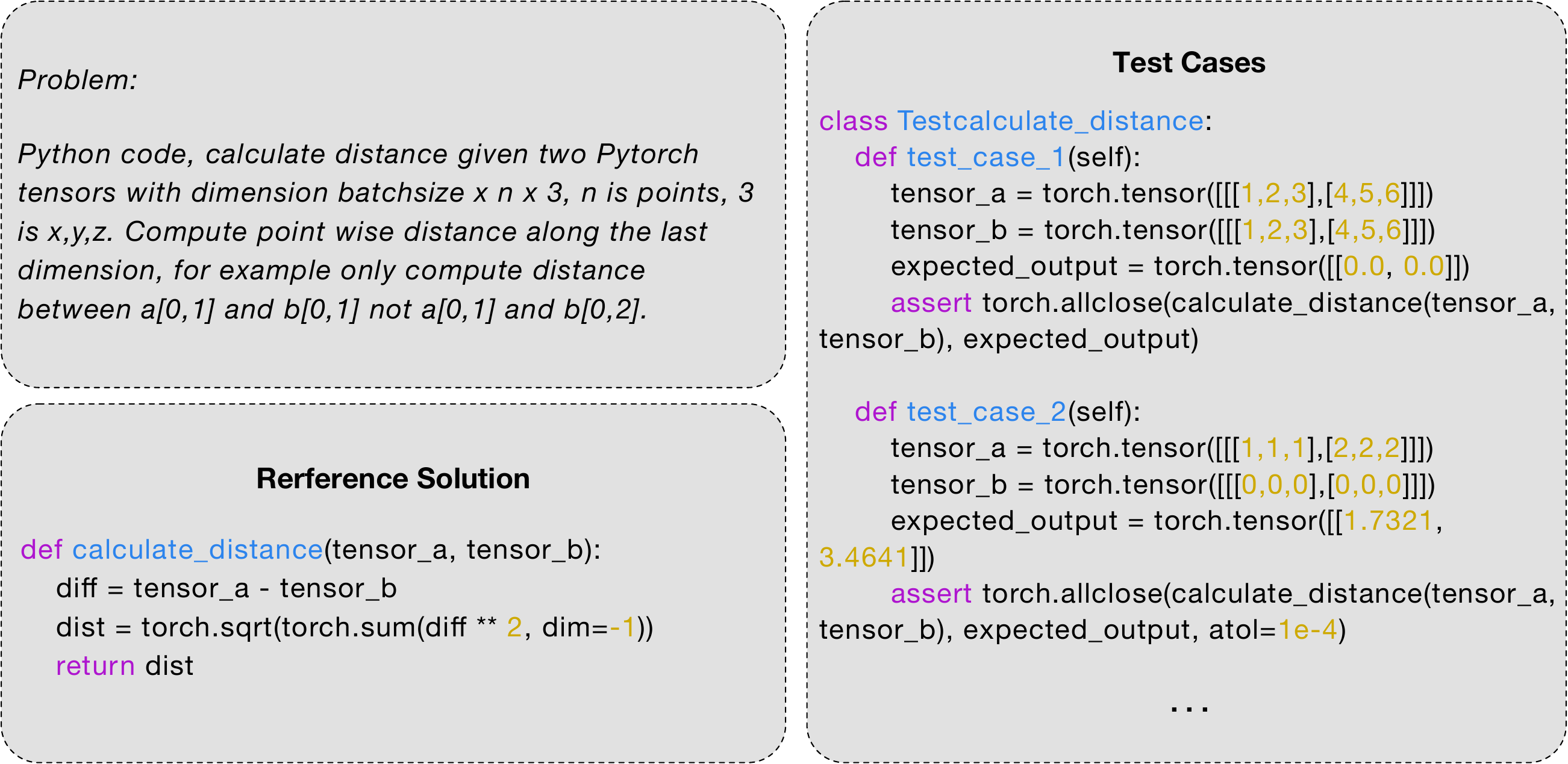}
    \caption{An example problem of Artificial Intelligence.}
    \label{fig:ai}
\end{figure*}

Figure~\ref{fig:ai} shows an example problem in artificial intelligence, requiring the calculation of the distance between each point of two tensors, where the dimension of each tensor is batchsize * n * 3, with the third dimension representing the coordinates of the points.

\begin{figure*}[h]
    \centering
    \includegraphics[width=0.95\textwidth]{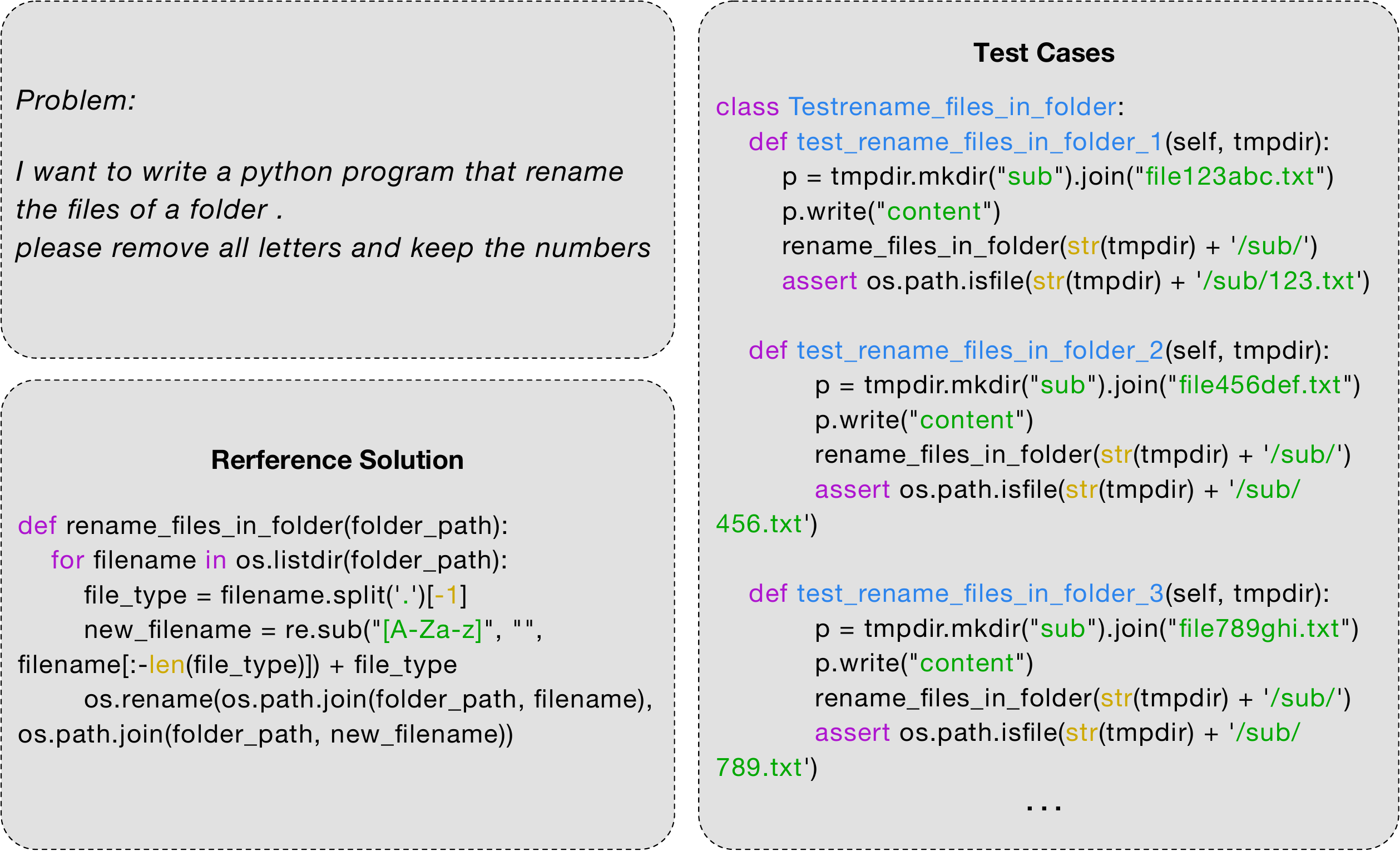}
    \caption{An example problem of System Administration.}
    \label{fig:os}
\end{figure*}

Figure~\ref{fig:os} presents an example problem in system administration, inquiring how to rename all the files within a folder according to a given rule.

\section{Extra Results}
\label{tab:ext-res}

Table~\ref{table:devset-result} shows the pass@1 results on the development set of \model. The results on the development set are essentially consistent with those on the test set, with some changes in the ranking among several models. This is due to differences in the distribution of problems across domains between the development set and the test set.

Table~\ref{tab:passk} shows the pass@k results of best-performing LLMs with each LLM family on \model, where $k \in \{10, 50\}$. We do not evaluate the performance on pass@k for ErnieBot4, CodeGeeX3, Claude-3, Gemini-1.5-Pro and Llama-3-Instruct due to limitations on the use of API and other resources.

\end{document}